# Super-Resolution for Remote Sensing Imagery via the Coupling of a Variational Model and Deep Learning

Jing Sun, Huanfeng Shen, Senior Member, IEEE, Qiangqiang Yuan, Member, IEEE, and Liangpei Zhang, Fellow, IEEE

***Abstract*—Image super-resolution (SR) is an effective way to enhance the spatial resolution and detail information of remote sensing images, to obtain a superior visual quality. As SR is severely ill-conditioned, effective image priors are necessary to regularize the solution space and generate the corresponding high-resolution (HR) image. In this paper, we propose a novel gradient-guided multi-frame super-resolution (MFSR) framework for remote sensing imagery reconstruction. The framework integrates a learned gradient prior as the regularization term into a model-based optimization method. Specifically, the local gradient regularization (LGR) prior is derived from the deep residual attention network (DRAN) through gradient profile transformation. The non-local total variation (NLTV) prior is characterized using the spatial structure similarity of the gradient patches with the maximum a posteriori (MAP) model. The modeled prior performs well in preserving edge smoothness and suppressing visual artifacts, while the learned prior is effective in enhancing sharp edges and recovering fine structures. By incorporating the two complementary priors into an adaptive norm based reconstruction framework, the mixed *L*1 and *L*2 regularization minimization problem is optimized to achieve the required HR remote sensing image. Extensive experimental results on remote sensing data demonstrate that the proposed method can produce visually pleasant images and is superior to several of the state-of-the-art SR algorithms in terms of the quantitative evaluation.**

This paragraph of the first footnote will contain the date on which you submitted your paper for review, which is populated by IEEE. This work was supported in part by the National Natural Science Foundation of China under Grant 42130108. (Corresponding author: Huanfeng Shen.)

Jing Sun is with the School of Resource and Environmental Sciences, Wuhan University, Wuhan 430079, China (e-mail: rainsunny@hotmail.com).

Huanfeng Shen is with the School of Resource and Environmental Sciences, Wuhan University, Wuhan 430079, China, and also with the Collaborative Innovation Center of Geospatial Technology, Wuhan University, Wuhan 430079, China (e-mail: shenhf@whu.edu.cn).

Qiangqiang Yuan are with the School of Geodesy and Geomatics, Wuhan University, Wuhan 430079, China (e-mail: yqiang86@gmail.com).

Liangpei Zhang is with the State Key Laboratory of Information Engineering in Surveying, Mapping, and Remote Sensing, Wuhan University, Wuhan 430079, China (e-mail: zlp62@whu.edu.cn).

## I. INTRODUCTION

WITH the development of remote sensing technology in recent years, the availability of remote sensing data has increased, and the need for remote sensing data with high spatial and temporal resolutions has become more and more urgent in geoscientific applications [1]. However, due to the limitation of the sensors and the complexity of the imaging environment, obtaining high-resolution (HR) remote sensing images often entails high data acquisition costs [2]. Therefore, remote sensing image super-resolution (SR) has been a major research focus, and is an effective way to enhance the spatial resolution of remote sensing images with an inexpensive and powerful solution. SR reconstruction technology aims to recover the HR image from one or multiple low-resolution (LR) images, to meet the needs of practical applications. However, the spatial resolution and clarity of remote sensing images are degraded due to the influence of the data acquisition and transmission processes [3], [4]. There have been many proposed works on the SR of remote sensing images employing different priors or network models. According to the number of input LR remote sensing images, the conventional SR approaches can be roughly categorized into single-frame super-resolution (SFSR) [5], [6], [7], [8] and multi-frame super-resolution (MFSR) [9], [10], [11], [12].

Compared to LR remote sensing images, HR images contain richer texture information and high-frequency details that can be lost in the process of data acquisition. The main challenge for remote sensing image SR is to recover the high-frequency details of the LR image, which are more sensitive to human perception. The early SFSR methods often employed a hand-crafted prior [13] to recover the HR remote sensing image from a single LR image. With the rapid development of deep learning, some studies [14], [15] have used remote sensing images to retrain a network designed for natural images. Due to the more complex structure of remote sensing images, compared to natural images, these methods are less effective than when applied to natural images. Subsequently, a large number of deep learning based methods [6], [16], [17] have emerged as a preferable choice for mining a generalizable prior and intra-image information from large-scale remote sensing data, and have achieved significant efficiency and generalization improvements in the SR reconstruction of remote sensing images.

Considering the unique characteristics of remote sensing images, the channel attention network was introduced by Yu *et*

*al*. [18] to aggregate sufficient feature information in each channel, and achieved a superior performance. Through experiments, it has been demonstrated that this strategy can effectively preserve the spatial detail information in the recovered image. In order to boost the generalization of SR models on various remote sensing scenes, a scene-adaptive strategy (MSAN) was employed by Zhang *et al*. [7] to accurately describe the structural characteristics of different scenes. In addition, according to the hierarchical distribution characteristics of remote sensing images, Xiao *et al*. [19] proposed a cross-scale hierarchical transformation method to effectively explore cross-scale representations in remote sensing images. More recently, a cross-sensor degradation modeling strategy was proposed for remote sensing image SR by Qiu *et al*. [20], which aims to bridge the gap between the images obtained by the source and target sensors. The above methods have utilized different structures for the cross-scale hierarchical characteristics and low prior information of remote sensing images, and the lack of detail in the HR space. However, the information used by the SFSR methods is limited to the spatial domain of the LR image, which does not consider the temporal information of remote sensing imagery. For remote sensing image processing, the performance of deep learning based SFSR is limited without the use of the complementary information between temporal frames.

On the other side, the MFSR approach can super-resolve the HR image by merging the temporal subpixel information from corresponding LR images, which are usually obtained from either geostationary orbit satellite or video satellite platforms. Since sequential satellite images contain complementary spatial and temporal information, the deep learning based MFSR methods need to simultaneously model the spatio-temporal relationship between sequential frames. To enhance the resolution of remote sensing images progressively, the progressively enhanced network for satellite image SR (PECNN) method composed of two subnetworks was proposed by Jiang *et al*. [21] to finely learn the structural information and low-level features in wide scenes. As for dealing with the noise in satellite images, a generative adversarial based edge-enhancement network was proposed by Jiang *et al*. [22] to enhance the high-frequency edge information in satellite video. While these methods have made progress in satellite image SR, the limited spatial information restricts their ability to reconstruct more precise textures. Differing from the previous MFSR methods for remote sensing images, the MVSRnet method [10] can merge the motion information among adjacent frames and highlight the importance of extracted features with an attention mechanism. This method utilizes optical flow estimation to warp frames based on the current frame and learns the progress of the multi-frame fusion from an external database. Salvetti *et al*. [23] proposed the residual attention MFSR network, which leverages feature extraction from multiple LR images of the same scene, resulting in reconstructed images with fine texture details. The experimental results demonstrated that this fusion approach can improve the quality of the reconstruction. However, the low efficiency of the spatio-temporal information fusion results in the poor generalization ability of the MFSR approach when applied to remote sensing images. More advanced methods learn the subpixel registration and achieve fusion simultaneously through a deep neural network. In order to model the spatio-temporal information collaboratively, a novel fusion strategy of temporal grouping projection and an accurate alignment module was proposed by Xiao *et al*. [24] for satellite video SR to effectively alleviate the alignment difficulties. In addition, Shen *et al*. [1] proposed an edge-guided SR (EGVSR) framework for satellite imagery, which can help the network focus more on the structure of ground objects and enrich the details in the SR results of remote sensing images. Most of the deep learning based methods typically simplify the image degradation model and use bicubic interpolation down-sampling to synthesize the LR images [8]. However, a model trained on simulated remote sensing images often lacks universality and generalizability when handling multiple degradations.

To handle the various unknown degradations in real-world remote sensing images, recent research has leveraged degradation estimation to reconstruct the SR image with a deep joint estimation network [25]. Nevertheless, degradation estimation methods are usually time-consuming and struggle to obtain an accurate estimation, resulting in poor SR performance due to the large estimation errors. Unlike the deep learning based methods, which are highly dependent on training samples, the variational model based approaches focus on designing some prior knowledge as the reconstruction constraint to regularize the super-resolved images without using any training samples [26]. These methods consider the imaging mechanism and construct the energy function according to a degradation model between the ideal image and the degraded observations, which is deterministic and theoretically reasonable [27]. Due to the rigorous theory, the variational model based methods are often more accurate than the traditional methods, and have the advantages of a strong noise reduction ability and convenient integration of spatial prior constraints.

The MFSR approach based on a variational model consists of a data fidelity term and a regularization term. Generally speaking, the data fidelity term measures the model error between the degraded observations and the ideal image, while the regularization term imposes the structural and statistical characteristics of the image itself as the model constraint to achieve a robust solution. A maximum *a posteriori* (MAP)-based MFSR method with $L1$ norm data fidelity and an edge-preserving Huber regularization prior was proposed by Shen *et al*. [28] to super-resolve multi-temporal Moderate Resolution Imaging Spectroradiometer (MODIS) images. Subsequently, several variational model based MFSR approaches [29], [30] have been proposed for remote sensing images captured with different satellite sensors and angles. Generally, the performance of the variational model based MFSR methods is highly dependent on the image priors, which define the different feature models of the images. Recently, the non-local self-similarity (NLSS) property of remote sensing images for natural scenes, where small textured regions tend to repeat multiple times at different locations inside the image, has become one of the most pervasive and powerful priors. In order to preserve fine texture details, Liu *et al*. [31] proposed a traditional SR framework that formulates a non-local regularization term with spatio-temporal domain NLSS. However, due to the high background complexity and large target scale variation of remote sensing images, although the traditional MFSR methods can improve the image resolution to a certain extent, they still have shortcomings in high-frequency edge detail recovery and noise artifact suppression.

Motivated by the need to recover the edge sharpness, many approaches [32], [33], [34], [35] have attempted to recover the high-frequency details by modeling and estimating the image edges and gradients. With the aim of imposing appropriate edge priors, Yang *et al*. [33] introduced a deep edge guided recurrent residual (DEGREE) network to reconstruct images progressively to the desired resolution. DEGREE further enhances the edge-preserving capability of the SR process, i.e., LR images and their edge plots can jointly infer the sharp edge details of the HR image during the recurrent recovery process. However, the edge priors contain only a small part of the high-frequency information, which limits the improvement in reconstruction performance. Clearly, the gradient prior is more effective in recovering sharp edges, since the lost high-frequency details such as edges and textures are mainly contained in the image gradient field. As the high-frequency information is mostly contained in the image gradient field, an effective SR method was proposed by Chen *et al*. [34] via learning-based gradient regularization and NLSS modeling. The gradient prior learned from the jointly optimized regression model was constructed as a gradient regularization term in the SR process. In addition, Song and Liu [35] trained a gradient prior learning network to regularize the image reconstruction and better recover the textural information and high-frequency components in the super-resolved image.

However, modeling the gradient field via a simple model ignores the local geometric structures of the gradients. Several SR algorithms [36], [37], [38] based on gradient profile representation have been developed to restore the edge sharpness of the reconstructed images. Since a sharp edge in an image is related to the gradient concentration perpendicular to the edge, the gradient transform is developed on the basis of statistical evaluation of the gradient profile fitting error. The LR gradient is converted to an HR gradient through statistical and parametric models, and then the transformed gradient field is taken as the constraint for HR image estimation [37]. Nevertheless, the parametric representation models are not flexible enough, using only a few parameters to capture the diverse transformation relationships between the LR and HR gradient profiles. As a result, the obtained HR images are usually over-sharpened or suffer from artifacts due to the uncertainty in the gradient field estimation. Thus, the utilization of a non-parametric transformation representation has been considered to construct a more adaptable gradient transformation model. Li *et al*. [38] designed an effective image SR approach by incorporating the gradient profile prior derived from example-based gradient field estimation, to enhance the edge sharpness and restore the image details in the super-resolved image. Although the learned gradient prior is more expressive than the parametric gradient profile models, the example-based architecture used in this framework has shortcomings in extracting the high-frequency features for SR reconstruction, especially in the complex detail areas.

On the one hand, these SR methods based on edge and gradient estimation are primarily tailored for natural images. Hence, in the context of remote sensing images, it is imperative to incorporate a gradient profile transformation model in the SR process. On the other hand, it is noteworthy that, due to the diverse target objects and significant changes in the high-frequency information distribution patterns within remote sensing data, such data often exhibit highly intricate spatial and hierarchical distributions in both the local and global features. Consequently, when designing the regularization term for remote sensing image SR, consideration should be given to incorporating both the local texture details and global structure information. According to the spatial distribution of the image features, image priors can be divided into two main categories: non-local priors and local priors. A non-local prior reveals the inter-scale and intra-scale redundant recurrence of small image patches, based on the self-similarity of the image patches. The non-local total variation (NLTV) prior is a promising and significant prior characterized by the internal spatial structure similarity of the image patches. A local prior can be formulated as various smoothness models based on the assumption of local smoothness. The smoothness models perform well in recovering smooth regions, but high-frequency details may be smoothed out. Due to the different emphases on the different characteristics of images, the non-local and local priors have their own merits and drawbacks in remote sensing image SR. By assembling multiple complementary priors, it is possible to make a trade-off between artifact reduction and detail preservation, suppressing the displeasing artifacts well and preserving the fine details.

Inspired by these works, we propose a gradient-guided MFSR method that employs a variational model with embedded learning based on local gradient regularization (LGR) and a non-local total variation (NLTV) prior. By exploiting both the learned LGR prior and the modeled NLTV prior, the proposed method benefits from the complementary properties of the learning-based and model-based SR approaches. More specifically, a deep residual attention network (DRAN) is trained to learn the horizontal and vertical gradients. The HR gradient profile is estimated from the learned gradient field via gradient profile transformation, which is used as the LGR constraint during the SR process to recover fine image details and enhance edge sharpness. Benefiting from the development of non-local means and cross-scale self-similarity methods, the internal spatial similarity of the gradient blocks is used to characterize the NLTV prior to suppress image noise and artifacts. In addition, to ensure the noise robustness and edge smoothness of the reconstructed image, a reliable and flexible adaptive fidelity norm model is established to suppress the artifacts that may be produced by the local gradient prior. By incorporating the two complementary priors into an adaptive norm based reconstruction framework, the HR estimate can be obtained via the alternating direction method of multipliers (ADMM) algorithm. The proposed method has the following benefits:

1) We propose to utilize the DRAN to learn the latent gradient field of the desired HR remote sensing image.

The learned gradient serves as an LGR prior in the SR process via gradient profile transformation.

2) The internal spatial similarity of gradient blocks is used to characterize the NLTV prior. This modeled prior is constructed as a regularization term to constrain the non-local features of the remote sensing imagery.
3) We propose an effective gradient-guided MFSR framework for remote sensing imagery by incorporating the learned LGR prior and the modeled NLTV prior. This framework combines the model-based and learning-based methods in a simple manner. The ADMM algorithm is exploited to effectively optimize the proposed minimization problem.
4) The proposed MFSR method takes advantage of the complementary modeled and learned priors, the artifacts in the super-resolved remote sensing image can be removed, and the fine structures and sharp edges can be well recovered. The extensive experiments validate the superior performance of the developed MFSR scheme.

The remainder of this paper is organized as follows. Section II introduces the proposed MFSR framework based on modeled and learned priors. In Section III, the experimental results are provided, including extensive comparisons and analyses. Finally, our conclusions are drawn in Section IV.

## II. Methodology

For the model-based MFSR methods, SR is usually considered as an ill-posed inverse problem defined by the reconstruction constraint and image prior. Clearly, the more accurate the model is, the better the SR performance will be. The image prior defines the different characteristic models of the images and play a pivotal role in the quality of a reconstructed HR image. In this section, we propose a gradient-guided MFSR method combining the model-based NLTV prior with the learning-based LGR prior to significantly improve the performance of remote sensing image SR. On the one hand, the initial HR gradient is learned from the DRAN. The final HR gradient profile with higher sharpness is then estimated from the learned initial gradient field via gradient profile transformation, and serves as the LGR constraint during the SR process. The novel LGR prior is good at fine detail recovery and edge sharpness enhancement. On the other hand, the NLTV prior is essentially a non-local and internal image based prior, which performs well in preserving edge smoothness and suppressing reconstruction artifacts. Benefiting from the complementary properties of the modeled and learned priors, HR remote sensing images are reconstructed with a better objective and subjective quality.

The model-based MAP method incorporates the two complementary prior constraints to estimate the desired HR image by minimizing an objective function of the posterior probability. In addition, to deal with the mixed noise and/or complicated model error in remote sensing image SR, a regularized framework with an adaptive fidelity norm is employed. Both the adaptive norm fidelity term and the joint regularization term guarantee the robustness of the proposed MFSR framework. The basic formula of the proposed MFSR method for remote sensing images is as follows:

$$\hat{z} = \arg\min_{z} \left\{ \underbrace{\sum_{k}^{K} \left\| y_k - DBM_k z \right\|_p^p}_{\text{fidelity term}} + \alpha \underbrace{P_{ext}(\nabla z)}_{\text{learned prior}} + \beta \underbrace{P_{int}(\nabla z)}_{\text{modeled prior}} \right\} \quad (1)$$

where $z$ is the ideal HR remote sensing image required to be reconstructed, and $p$ is the fidelity norm value in the interval $[1,2]$ . $y_k$ is the $k$-th observed LR image, where $k = 1,2,\cdots,K$ , with $K$ being the number of LR images. $B$ represents the blur matrix, including the sensor blur, optical blur, and atmospheric turbulence. $M_k$ is the motion matrix, and the relative movement parameters can be estimated using the subpixel shifts of the multiple LR images. $D$ denotes the down-sampling matrix. $\alpha$ and $\beta$ represent the regularization parameters, which provide a trade-off between the data fidelity term and the regularization term. The fidelity term measures the reconstruction error to ensure that pixels in the reconstructed HR image are close to the real values. The model-based and learning-based priors are combined with model features (such as non-local similarity) and learned features (such as the learned gradient features) to regularize the solution space of the HR image.

The flowchart of the proposed gradient-guided MFSR method is demonstrated in Fig. 1. This framework contains three main parts: 1) learning-based gradient estimation and LGR term construction through gradient profile transformation; 2) non-uniform interpolation (NUI) initialization and NLTV regularization term construction; and 3) energy function optimization and HR image updating. The details of these parts are introduced in the following subsections.

### A. Gradient-Learning Network

The main parts of remote sensing images belong to the low-frequency components, which are a comprehensive measure of the image intensity. Meanwhile, the high-frequency components correspond to the parts of the image that change dramatically, i.e., the edges or noise and details of the image. Aimed at better reconstructing the texture information and high-frequency components of remote sensing images, the effective integration of subpixel complementary information between image sequences and the reliable prediction of the high-frequency components should be considered simultaneously. To achieve an HR remote sensing image with more fine details and sharp edges, we introduce an external gradient prior as a regularization constraint to recover the high-frequency details. An effective gradient regularization prior needs to be established on accurate gradient information in the SR reconstruction process, but the gradient of the LR image will not have sharp details, so it is necessary to use the existing gradient information to reconstruct the missing gradient information.

Differing from the variational model based approaches, the deep learning based methods introduce external training datasets to capture image features, without the need for explicit expression of the regularization term. In addition, the deep learning based approaches are more capable of information extraction and feature fusion with an end-to-end learning framework. Therefore, considering the excellent learning capacity of convolutional neural networks, the DRAN is constructed and trained for latent gradient prior learning. As shown in Fig. 2, the gradient-mapping network contains a spatial feature fusion module (SFFM), one upsampling layer, and several convolutional layers. The attention mechanism is applied in the SFFM, which consists of three spatial attention modules (SAMs), one residual dense module (RDM), and a convolutional layer.

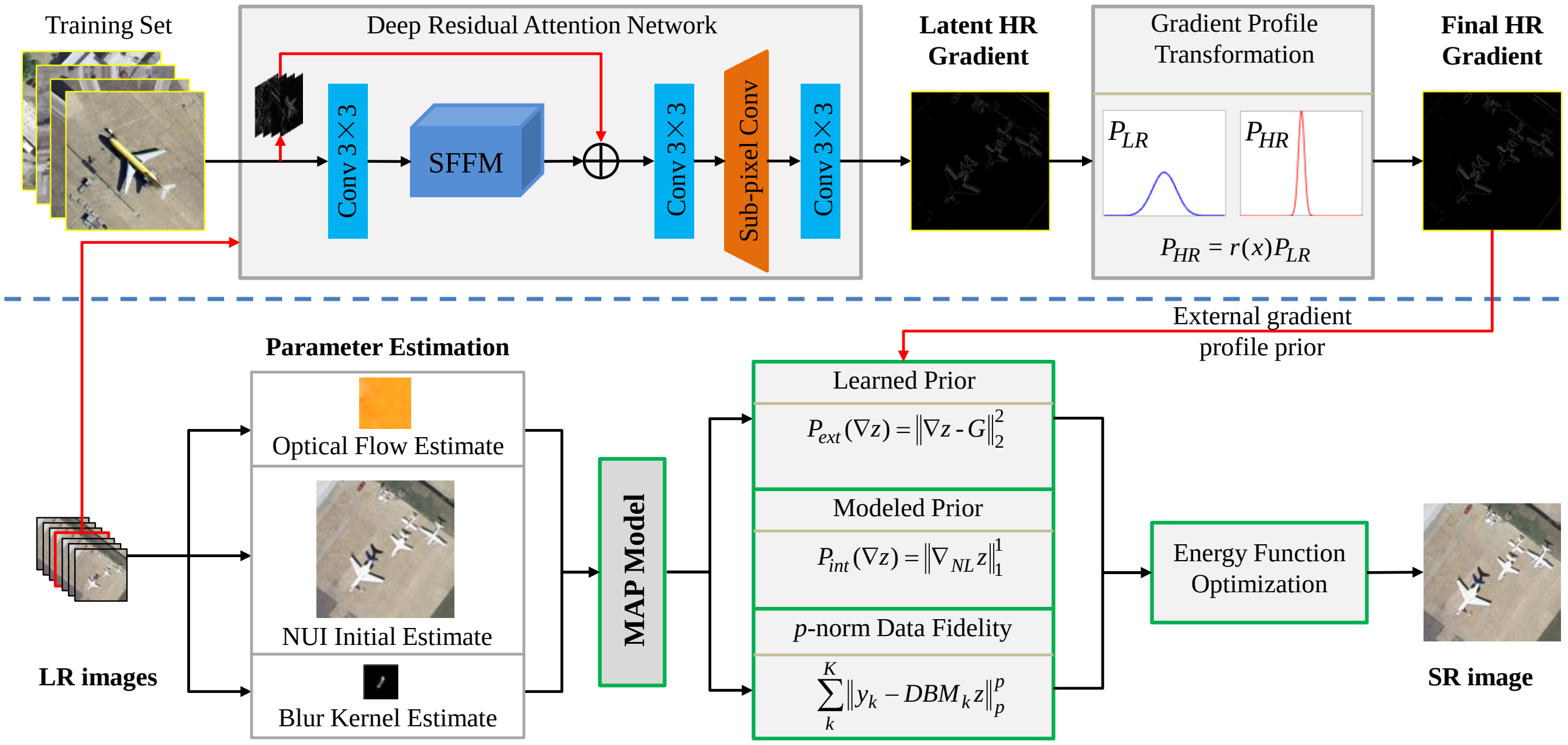

Fig. 1. Flowchart of the gradient-guided MFSR algorithm.

In the proposed framework, the DRAN is used to predict a latent HR gradient field that has accurate and clear contours. Due to the high-frequency characteristics of image gradients, the mapping from LR gradient to HR gradient is actually high-frequency to high-frequency mapping. During image degradation, the high-frequency components are destroyed and are more unstable than the low-frequency components. Therefore, learning the HR gradient prior directly from the LR gradient is difficult for the network as only the high-frequency information is given. Conversely, it is easier to learn from the LR image itself, which can also provide reliable low-frequency information. In the proposed approach, the proposed DRAN takes the LR image and the corresponding horizontal and vertical gradients as input.

The discrete gradient operators $[-1/2,\ 0,\ 1/2]$ and $[-1/2,\ 0,\ 1/2]^T$ are separately employed to extract the horizontal gradient $g_y^h$ and vertical $g_y^v$ gradient of the LR image $y$, respectively. Note that the binary strategy is eliminated to avoid the appearance of false edges and missing image features. The LR image $y$ and the gradient $g_y^h$, $g_y^v$ are then taken as the input of the network. The shallow feature $F_0$ is first extracted from the input LR image $y$ and can be formulated as $F_0 = f_{\mathrm{Conv3}}(y)$, where $f_{\mathrm{Conv3}}$ denotes a convolution operation with the kernel size of 3 × 3.

*1) Spatial Attention Module:* The SAM is able to keep the good properties of the original input features, not only emphasizing the more important features, but also suppressing the less useful features. Since the high-frequency features are more important for the HR gradient reconstruction, the spatial attention mechanism is employed to further enhance the gradient information. The structure of the SAM is displayed in Fig. 3, which consists of one residual block (RB) and a spatial attention block (SAB).

The initial feature $F_0$ is fed into a stack of SAMs to make full use of the shallow layer information and capture the long-range dependencies. The deep feature map of the $n$-th SAM can be expressed as follows:

$$F_{S_n} = f_{S_n}\left(F_{S_{n-1}}\right) + W_S F_0 \tag{2}$$

where $f_{S_n}$ denotes the function of the $n$-th SAM. $F_{S_{n-1}}$ and $F_{S_n}$ represent the input and output feature maps of the $n$-th SAM, respectively. $W_S$ is a learnable parameter, and "+" is the element-wise addition operation.

In the SAM, the initial feature $F_0$ is first processed by an RB that consists of two convolutional layers with the kernel size of 3 × 3 and a parametric rectified linear unit (PReLU) activation function [39]. The output feature of the RB can be written as follows:

$$F_{R_n} = f_{Conv3}\left(\rho\left(f_{Conv3}\left(F_{S_{n-1}}\right)\right)\right) \tag{3}$$

where $\rho(\cdot)$ denotes the PReLU activation function, and $F_{R_n}$ represents the intermediate feature of the $n$-th SAM. The obtained intermediate feature $F_{R_n}$ is then fed into a SAB, which uses the spatial attention mechanism to make the model focus on the high-frequency region of the image.

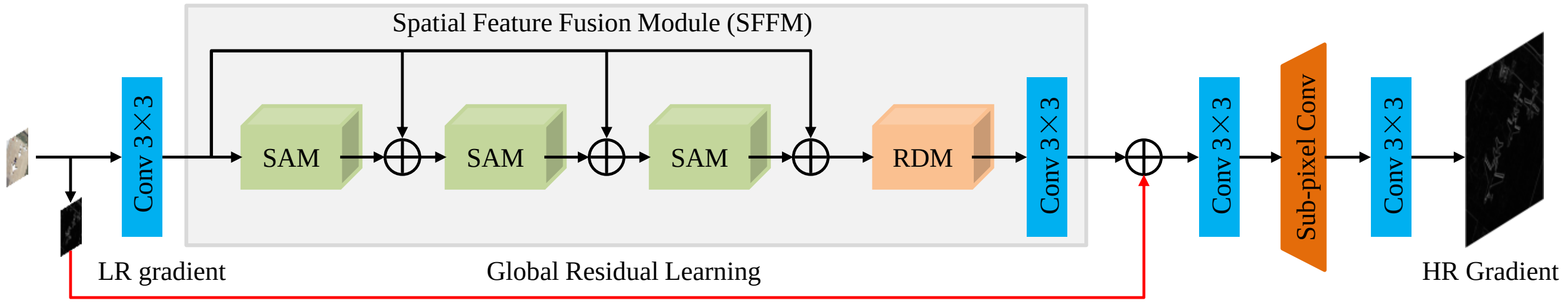

Fig. 2. Architecture of the gradient-mapping DRAN.

In the SAB, both maximum and average pooling operations are employed simultaneously to squeeze the spatial information of the input feature map. The weights of the spatial attention mechanism can be expressed as follows:

$$W_{SA}=\sigma\left(f_{Conv3}\left(\rho\left(f_{Conv3}\left(\left[P_{avg}(F_{R_n});P_{max}(F_{R_n})\right]\right)\right)\right)\right) \quad (4)$$

where $[\cdot;\cdot]$ represents the concatenation operation; $P_{avg}$ and $P_{max}$ denote the average pooling operation and the maximum pooling operation, respectively; and $\sigma(\cdot)$ represents the sigmoid activation function. The attention map is then used to modulate the intermediate features. The modulated features $F_{S_n}$ can be obtained by multiplying the intermediate features $F_{R_n}$ with the spatial weight map $W_{SA}$, which can be written as follows:

$$F_{S_n}=F_{R_n}\otimes W_{SA} \quad (5)$$

where $\otimes$ denotes element-wise multiplication.

Finally, the modulated features $F_{S_N}$ of the last SAB and the initial feature $F_0$ are connected by a skip connection structure to generate the final output feature $F_S$ of the last SAM.

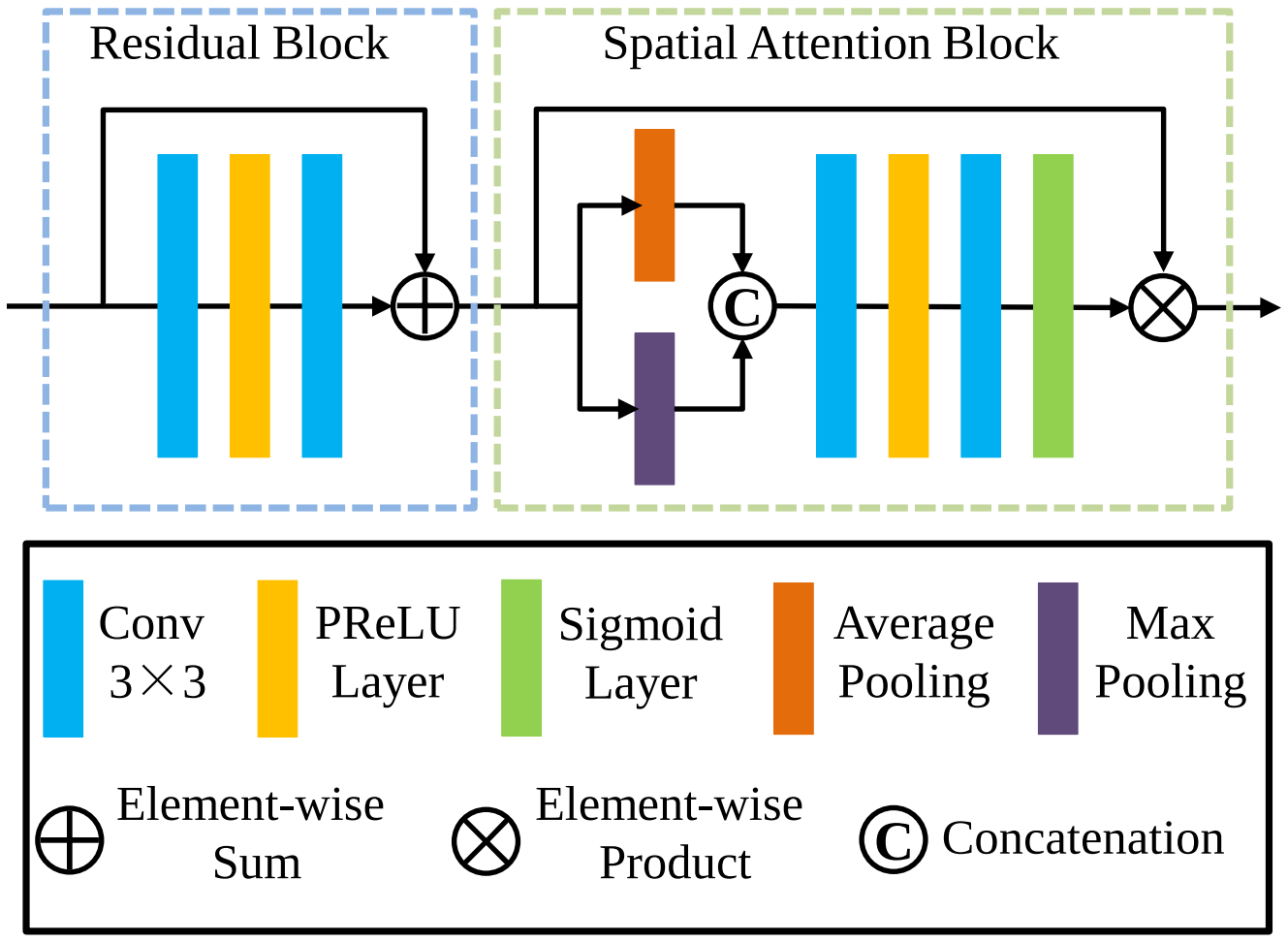

Fig. 3. Architecture of the spatial attention module (SAM).

*2) Residual Dense Module:* To fully utilize the hierarchical features, the final enhanced feature $F_S$ from the SAM is then passed into an RDM. The RDM combines the residual structure and dense connections to produce improved features and encourage feature reuse, as shown in Fig. 4.

In the RDM, each layer aggregates the feature maps from all the preceding layers. More specifically, to generate the feature maps effectively, the input for each convolutional layer is the concatenation of the outputs from all the previous convolutional layers. A $1\times1$ convolutional layer $f_{Conv1}(\cdot)$ is used for feature pooling and dimension reduction after the concatenated feature map. Thus, the concatenated feature map is merged along the channel dimension to produce the merged feature map $F_{C_m}$. The concatenate module can be written as follows:

$$F_{C_m}=f_{Conv1}\left(f_{Concat}\left(F_{D_1},F_{D_2},\cdots,F_{D_m}\right)\right) \quad (6)$$

$$F_{D_{m+1}}=\rho\left(f_{Conv3}\left(F_{C_m}\right)\right) \quad (7)$$

where $F_{C_m}$ denotes the output feature maps of the $m$-th concatenation operation $f_{Concat}$. $F_{D_m}$ and $F_{D_{m+1}}$ represent the input and output feature maps of the $(m+1)$-th convolutional layer, respectively. Finally, the merged features $F_{D_m}$ containing more representational information can be obtained through the RDM.

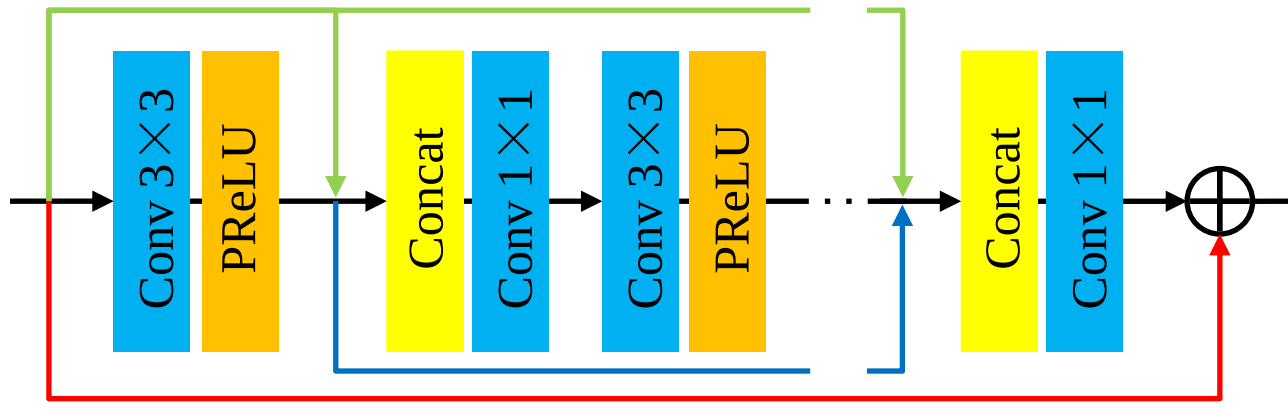

Fig. 4. Architecture of the residual dense module (RDM).

*3) Residual Reconstruction Module:* The residual reconstruction module mainly consists of one upsampling layer and several convolutional layers. The output feature of the last RDM is passed through a convolutional layer, followed by global residual learning, which is aimed at facilitating the feature representation and preserving the base performance of the network. As stated before, the extracted gradient of the LR image $y$ is denoted as $g=\left[g_y^h, g_y^v\right]$, which is connected with the final merged feature $F_D$ by the global residual learning. The output of the global residual learning is denoted as $F_G$, which is formulated as follows:

$$F_G = f_{Conv3}(F_D) + g \tag{8}$$

In the proposed approach, the common subpixel convolutional layer is employed as the post-upsampling layer. At the end of the network, the final latent HR gradient $\hat{g} = \left[ \hat{g}_y^h, \hat{g}_y^v \right]$ can be obtained through the following formula:

$$\hat{g} = f_{Conv3}\left( f_{Up}\left( f_{Conv3}\left(F_G\right)\right)\right) \tag{9}$$

where $f_{Up}(\cdot)$ represents the subpixel convolutional layer.

Finally, the end-to-end gradient-mapping network is optimized with a certain loss function. An $L1$ regularization term is constructed to constrain the training of the gradient-learning network, so that it can generate a gradient field with more accurate details and provide a gradient prior for the SR process. The gradient-learning loss function is defined as follows:

$$Loss = \frac{1}{N}\sum_{i=1}^{N}\left\| f_{DRAN}(y_i, g_{y_i}; \theta) - g_{z_i} \right\|_1^1 \tag{10}$$

where $f_{DRAN}(\cdot)$ denotes the whole end-to-end mapping function, and $\theta$ represents the parameter of the network.

The proposed network has 29 convolutional layers. Since the DRAN mainly transforms the spatial distribution of edges and textures in the LR and HR gradient maps, the designed lightweight gradient-learning network can capture the structural dependency and generate accurate gradient maps.

### *B. Gradient Profile Transformation*

Since gradients reveal the local variations in image intensity, we chose to learn the gradient-mapping function from LR image to HR image via the above DRAN. However, modeling the gradient field via the lightweight gradient-learning network only considers the marginal distributions and ignores the local geometric structures of the gradients. In this subsection, we describe how the deep learning based gradient-mapping network is combined with gradient profile transformation to estimate the HR gradient profiles, which further make up an HR gradient image to provide the gradient field prior for the proposed MFSR method.

The gradient spatial information determines the texture details and edge information of an image. The gradient profile transformation approach is introduced to enhance the useful gradient spatial information and reconstruct the HR gradient image with more accurate and sharp details. The gradient profile is a feature describing the spatial layouts of the edge gradients, and is defined as a 1D profile of the gradient magnitudes along the gradient direction of an edge pixel [37]. In brief, the gradient profile prior is a parametric distribution that describes the shape and sharpness of the image gradients.

The generalized Gaussian distribution (GGD) model is applied to model the normalized gradient profile, and is defined as follows:

$$h(x;\sigma,\lambda) = \frac{\lambda\alpha(\lambda)}{2\sigma\Gamma(\frac{1}{\lambda})}\exp\left\{ -\left[ \alpha(\lambda)\left|\frac{x}{\sigma}\right| \right]^{\lambda} \right\} \tag{11}$$

where $\Gamma(\cdot)$ is the gamma function, and $\sigma$ represents the sharpness of the gradient profile. $\alpha(\lambda) = \sqrt{\Gamma(3/\lambda)/\Gamma(1/\lambda)}$ is the scaling factor which makes the second moment of GGD equal to $\sigma^2$. $\lambda$ is the shape parameter which controls the overall shape of the distribution. As indicated in [37], the value of $\lambda = 1.6$ represents a suitable generic model for the gradient profiles sharpness and is also independent of the image resolution. Based on this property, it can be deduced that the profile sharpness is the only parameter that necessitates further investigation.

The HR gradient field is estimated by transforming the LR gradient field with the gradient profile model. Intuitively, the HR gradient profile $p_{HR}$ can be derived through multiplying the LR gradient profile $p_{LR}$ by the transform ratio $r(d)$. The ratio between the two gradient profiles is computed as follows:

$$\begin{aligned} r(d) &= \frac{h(d;\sigma_{HR},\lambda)}{h(d;\sigma_{LR},\lambda)} \\ &= \frac{\sigma_{LR}}{\sigma_{HR}}\cdot\exp\left\{ -\left(\frac{\alpha(\lambda)\cdot d}{\sigma_{HR}}\right)^{\lambda} + \left(\frac{\alpha(\lambda)\cdot d}{\sigma_{LR}}\right)^{\lambda} \right\} \end{aligned} \tag{12}$$

where $d$ is the curve distance to the edge pixel along the gradient profile. Sun *et al.* [36] suggested a function for sharpness enhancement: $\sigma_{HR} = \sigma_{LR}(1 - e^{-\mu\sigma_{LR}})$.

Subsequently, gradient profile transformation is performed on the learned gradient field, based on the gradient profile sharpness mapping between the LR image and the HR image, in order to obtain the final HR gradients. Consequently, the final HR gradient field $G(x)$ is directly estimated from the learned gradient field $\hat{g}(x)$ by the following formula:

$$G(x) = r(x)\cdot\hat{g}(x) \tag{13}$$

The gradient field transformation is performed for each pixel rather than each gradient profile. The reason for this is that the new gradient is required at each pixel grid, while a pixel grid may not fit the gradient profile exactly. If each gradient profile is transformed, many subpixel gradient values will be generated, and it will be difficult to determine the gradient magnitude from the subpixel gradient. In contrast, the gradient field transformation for each pixel is more straightforward and easy to implement. Finally, the estimated HR gradient is taken as a gradient constraint in the subsequent image reconstruction to produce a better HR image.

### *C. Adaptive Norm Reconstruction Model*

In the proposed MFSR approach, the fidelity term guarantees that the solution accords with the degradation process of remote sensing images, while the regularization term enforces the desired property of the output HR image. In this subsection, we introduce the proposed MFSR algorithm in detail. In the traditional MFSR methods, the initial HR estimation is obtained by performing bicubic or bilinear interpolation on the reference LR image. However, if the quality of the reference LR image is poor, the quality of the initial HR estimation will be unsatisfactory. Based on MAP estimation, we exploit NUI initialization to obtain the initial HR estimation for the image SR.

*1) Adaptive Norm Data Fidelity:* The fidelity term is mainly responsible for dealing with motion outliers, image noise, and model errors. For the complicated types of noise and model error in remote sensing image SR, both the $L1$ norm and $L2$ norm have their advantages and drawbacks. To deal with different types of noise and blur, we employ adaptive $p$-norm data fidelity [40] to select the error norm adaptively, instead of using fixed $L1$ and $L2$ norms. The adaptive $p$-norm data fidelity is formulated as follows:

$$D(y,z)=\sum_{k}^{K}\left\|y_k-DBM_k z\right\|_p^p \tag{14}$$

The norm value of $p$ reflects the proportional relationship of the different distributions in the mixed error model. According to the distribution models of different types of noise, the optimal $p$-norm is determined adaptively, based on a generalized likelihood ratio test (GLRT), which is formulated as follows:

$$p=\begin{cases}1 & 0<\gamma\le 0.112\\ a\tan(b\gamma+c)+d & 0.112<\gamma\le 0.798\\ 2 & 0.798<\gamma<1\end{cases} \tag{15}$$

where $\gamma=\sigma_L/\sigma_G$ is the ratio between the Gaussian and Laplacian distributions. $\sigma_G$ and $\sigma_L$ are the variance of the Gaussian noise and impulse noise distribution, respectively. As described in [40], the model parameters *a*, *b*, *c*, and *d* are obtained from a large number of experimental statistics.

For each remote sensing image, the noise vector is computed according to the image observation model. The two variances $\sigma_G$ and $\sigma_L$ of the noise can then be computed to obtain the corresponding $\gamma$ value. According to (15), we can adaptively determine the optimal $p$-norm value for each test image. For the $p$-norm model, the Lagrange approximation method can be flexibly applied to solve it.

*2) Reconstruction Model With Combined Prior:* Given that SR is regarded as an ill-posed minimization problem, it is essential to implement a specific regularization to constrain the solution space. In the proposed approach, we construct the reconstruction framework by coupling the NLTV prior and the LGR prior, which respectively consist of the reconstruction constraint in the image domain and the gradient domain.

Benefiting from the development of NLSS methods, we integrate the non-local property into the algorithm framework by considering the spatial similarity of the gradient patches. The non-local similarity information can be obtained by clustering the similar patterns throughout a remote sensing image, from which a non-local regularization term is formulated to enhance the texture details of the reconstructed image. Mathematically, we let $z_i$ denote the image patch at location $i$, and the non-local gradient operator at the pixel location is defined as follows:

$$\nabla_{NL} z=\sum_{i\in\Omega(z)}\sum_{j\in\Omega(i)}\left(\left(z_j-z_i\right)\sqrt{w_{i,j}}\right) \tag{16}$$

where $\Omega(z)$ is the index set for all the pixels of $z$, and $\Omega(i)$ denotes the index set for similar patches of $z_i$. $w_{i,j}$ represents the similarity weight between patches $z_i$ and $z_j$, which is defined by:

$$w_{i,j}=\exp\left(-\frac{G\left(z_j-z_i\right)}{2\eta^2}\right) \tag{17}$$

where $\eta$ is a parameter set to the standard deviation of the image noise. $G(\cdot)$ is Gaussian filter function to reduce the influence of noise on the weight. To effectively exploit the non-local redundancy within the remote sensing imagery, the NLTV prior is formulated as follows:

$$P_{int}(\nabla z)=\left\|\nabla_{NL} z\right\|_1^1 \tag{18}$$

In the variational embedded learning MFSR method, the NLTV regularization term is exploited to enhance the edge information and suppress artifacts in the smooth regions of the super-resolved image. The other prior is the LGR prior, derived from the gradient profile transformation, which considers the distribution of the image gradients along local image structures. The gradient constraint requires that the gradient field of the recovered HR image should be consistent with the transformed HR gradient field. Thus, the LGR prior can be formulated by:

$$P_{ext}(\nabla z)=\left\|\nabla z-G\right\|_2^2 \tag{19}$$

Using this LGR constraint, the gradient profile of the reconstructed HR image is encouraged to have the desired statistics learned from the training images. By coupling the non-local reconstruction constraint and local gradient domain constraint in the MAP framework, the performance of the gradient-guided MFSR method is further improved. Inserting (18) and (19) into (1), the objective function of the proposed gradient-guided MFSR method can be rewritten as:

$$\hat{z}=\arg\min_{z}\left\{\sum_{k}^{K}\left\|y_k-DBM_k z\right\|_p^p+\alpha\left\|\nabla z-G\right\|_2^2+\beta\left\|\nabla_{NL} z\right\|_1^1\right\} \tag{20}$$

In the MAP-based MFSR algorithm, the adaptive norm fidelity term is exploited to ensure robustness in dealing with different noise types. The combined prior is used to enhance the edge information of the image and suppress the artifacts in the smooth regions. Typically, the geometric registration and the blur can be estimated from the input data, and used with the SR model to reconstruct the super-resolved image. In this work, the warping matrix $M$ and blur matrix $B$ are computed with the optical flow approach [41] and the blind blur kernel estimation method [42], respectively. Finally, the combined SR problem is solved via the ADMM algorithm.

*3) Energy Function Optimization:* Since $p$ is an arbitrary value in an interval, the conventional linear optimization methods cannot be employed directly. In the proposed approach, the iteratively reweighted norm (IRN) method [43] is employed for the linearization of the fidelity term. The main idea of the IRN algorithm is to transform the indefinite norm problem into an equivalent $L2$ norm problem by selecting a suitable weight coefficient matrix, so that the objective function can be optimized by a conventional solution method. In order to simplify the problem, $DBM_k$ can be regarded as a system matrix $A_k$ . Introducing the idea into the energy function in (20), at iteration $n+1$, the solution $\hat{z}^{n+1}$ can be obtained by:

$$\hat{z}^{n+1} = \arg\min_{z} \left\{ \begin{array}{l} \sum_{k}^{K} \left\| W_{z^n}^{1/2} \left( y_k - A_k z \right) \right\|_2^2 + \alpha \left\| \nabla z - G \right\|_2^2 \\ + \beta \left\| \nabla_{NL} z \right\|_1^1 \end{array} \right\} \quad (21)$$

where $W_{z^n}^{1/2}$ is the weight coefficient matrix, which is defined as follows:

$$W_{z^n}^{1/2} = diag\left( \varphi_\varepsilon \left( y_k - A_k z^n \right) \right) \quad (22)$$

with

$$\varphi_\varepsilon(x) = \begin{cases} |x|^{p-2} & if\ |x| > \varepsilon \\ \varepsilon^{p-2} & else \end{cases} \quad (23)$$

where $\varepsilon$ is a small positive number to guarantee global convergence, which is fixed as $10^{-5}$ in the proposed approach.

Based on the variable splitting minimization approach, the ADMM algorithm [44] is adopted to solve the hybrid $L1$- and $L2$-regularized minimization problem in (21). The auxiliary variable $b$ representing $z$ is introduced to decouple the $L1$ and $L2$ parts. The objective function can be rewritten as:

$$\hat{z}^{n+1} = \arg\min_{z} \left\{ \begin{array}{l} \sum_{k}^{K} \left\| W_{z^n}^{1/2} \left( y_k - A_k z \right) \right\|_2^2 + \\ \alpha \left\| \nabla z - G \right\|_2^2 + \beta \left\| \nabla_{NL} b \right\|_1^1 \end{array} \right\} \quad s.t.\ b = z \quad (24)$$

By transforming (24) to generate an unconstrained problem with the augmented Lagrangian algorithm, it can be rewritten as follows:

$$\hat{z}^{n+1} = \arg\min_{z} \left\{ \begin{array}{l} \sum_{k}^{K} \left\| W_{z^n}^{1/2} \left( y_k - A_k z \right) \right\|_2^2 + \alpha \left\| \nabla z - G \right\|_2^2 \\ + \tau u^T \left( z - b \right) + \frac{\tau}{2} \left\| z - b \right\|_2^2 + \beta \left\| \nabla_{NL} b \right\|_1^1 \end{array} \right\} \quad (25)$$

where $\tau$ is a penalty parameter, and $u$ is a Lagrangian multiplier. To tackle the optimal minimization problem in (25), an iterative algorithm is designed to alternately minimize $z$, $b$, and $u$ independently, by fixing the other variables. The sub-problem optimization can be rewritten as follows:

$$z^{n+1} = \arg\min_{z} \left\{ \begin{array}{l} \sum_{k}^{K} \left\| W_{z^n}^{1/2} \left( y_k - A_k z \right) \right\|_2^2 + \alpha \left\| \nabla z - G \right\|_2^2 \\ + \tau (u^n)^T \left( z - b^n \right) + \frac{\tau}{2} \left\| z - b^n \right\|_2^2 \end{array} \right\} \quad (26)$$

$$b^{n+1} = \arg\min_{z} \left\{ \beta \left\| \nabla_{NL} b \right\|_1^1 + \tau u^T \left( z - b \right) + \frac{\tau}{2} \left\| z - b \right\|_2^2 \right\} \quad (27)$$

$$u^{n+1} = u^n + z^{n+1} - b^{n+1} \quad (28)$$

For the $z$ sub-problem in (26), a preconditioned conjugate gradient method is used to quickly find the optimal value. The solution of $b$ in (27) is obtained using a simple shrinkage operation based on [45]. According to (28), $u$ can then be determined directly from the given variables $z$ and $b$ . By iteratively optimizing $z$ , $b$ , and $u$ , the optimal solution of (21) can be obtained in a fast and stable manner.

## III. Experimental Results

The performance of the proposed gradient-guided MFSR method was evaluated in several experiments. Initially, the proposed method was applied on the test datasets and compared with some of the state-of-the-art SR approaches. At the same time, simulation experiments on noisy images were further conducted to verify the robustness of the proposed method to noise. In addition, the effectiveness of the two incorporated regularization terms was validated using different extended versions of the proposed method. The detailed steps are presented in the following sections.

### *A. Experimental Setup*

*1) Degradation Models:* In the experiments, synthetic data with the ground truth were used to quantitatively analyze the proposed method, as well as make a fair comparison with the other methods at a scale factor of 4×. For each HR image from the test sets, we generated a set of K = 16 images with different subpixel shifts applied before further degradation. The 16 HR images were then blurred by a 3 × 3 isotropic Gaussian kernel with standard deviation 1. Finally, the row and column of the blurred image were down-sampled by a factor of 4. Additionally, in the robustness analysis described in Section III-C, the LR images were further contaminated by additive white Gaussian noise with a variance of 0.005.

In the MFSR method, the central frame of the LR sequence was chosen as the reference frame. The regularization parameters $\alpha$ and $\beta$ were determined empirically based on numerous experiments to produce the best performance. Since minimizing the objective function by the preconditioned conjugate gradient method usually converges within 30 iterations, the maximum iteration number was set to 30 in the proposed method.

*2) Dataset and Training Settings:* We chose the UC Merced dataset [46], which is a collection of remote sensing images with a relatively high spatial resolution (0.3 m/pixel), to evaluate the proposed method. As shown in Fig. 5, the UC Merced dataset contains 21 different scene categories, with each class comprising 100 images of 256 × 256 in the RGB space. The UC Merced dataset was used as the experimental data and divided into two disjoint parts. The first part contained images in the ten categories of airplane, baseball diamond, buildings, dense residential, freeway, harbor, intersection, mobile home park, storage tanks, and tennis court. We first selected 900 images from the ten categories, made up of 90 images for each class, to build the training dataset. The remaining 100 images formed test set I. The second part included images in the remaining 11 classes of agricultural, beach, chaparral, forest, golf course, medium residential, overpass, parking lot, river, runway, and sparse residential, which made up test set II. The reason for this division was to measure the generalization ability of the proposed method. The images in test set I were similar to those in the training set, while the images in test set II were quite different from those in the training set. Therefore, the performance under these test sets could better validate the generalization ability of the proposed method.

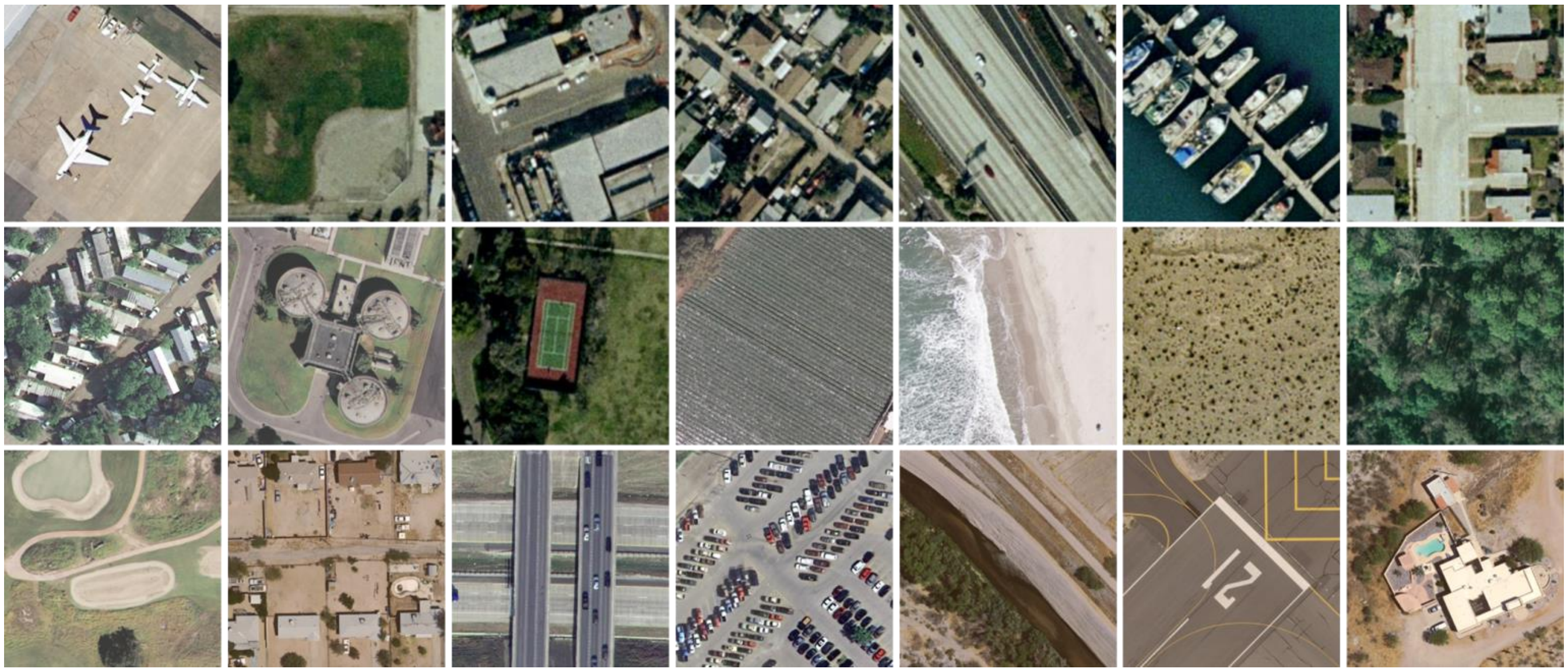

Fig. 5. Examples of the 21 different scene categories in the UC Merced dataset.

In the DRAN training phase, we exploited the above degradation models without subpixel shifts to simulate the LR input images. In total, 10% of the training samples were randomly selected as the validation set for the model selection, and the other 90% were used for the training. The factor of the residual scaling was set to 0.2. The weights of the network were initialized based on the method in [39]. In each mini-batch, 32 degraded LR images with a patch size of 48 × 48 were provided as inputs for the model, and the corresponding HR image served as the ground truth for calculating the loss. The models were optimized using the ADAM optimizer [47] with momentum parameters $\beta 1 = 0.9$, $\beta 2 = 0.999$, and $\varepsilon = 10^{-8}$. The initial learning rate was set to $10^{-4}$ and then decreased by half every 10 epochs. A total of 100 epochs were used for training the models since more epochs did not bring further improvements. All the experiments were implemented using the Caffe framework and MATLAB R2022a on an Nvidia RTX GPU.

*3) Comparison Baselines:* To comprehensively compare the proposed method with the state-of-the-art SR methods and classical SR methods, the comparison baseline methods were made up of bicubic interpolation, two deep learning based video SR methods (the recurrent back-projection network (RBPN) [48] and BasicVSR [49]), a spatially weighted bilateral total variation method (SWBTV) [50], and a joint prior based method (L0RIG) [51]. For the two deep learning based approaches, we used the training data to properly retrain these models with the specific training settings in their corresponding articles. Since the human visual system is more sensitive to the luminance component than the chrominance components, we converted the color RGB frames to the YCbCr color space and reconstructed only the luminance component with the proposed algorithm. Bicubic interpolation was used for the other components.

*4) Evaluation Metrics:* Image enhancement or visual quality improvement can be subjective because the perception of better image quality can vary from person to person. To assess the image quality of the SR reconstructed results, two classical evaluation criteria—the peak signal-to-noise ratio (PSNR/dB) and the structural similarity index measure (SSIM)—were chosen to measure the performance of the different SR methods [52]. The higher the quantitative measure, the better the quality of the reconstructed image. Since no ground-truth HR image was available for the experiments on the real sequences, we introduced no-reference image evaluation metrics—the natural image quality evaluator (NIQE) [53] and the perception-based image quality evaluator (PIQE) [54]—to further evaluate the quality of the real image SR results. Smaller values of NIQE and PIQE indicate better SR results. In addition, to give a fair comparison, we cropped the image boundary pixels before evaluation. The best results in Table I are highlighted in red/bold, while the second-best results are in green/underlined.

## *B. Synthetic Data Experiments*

In this section, both the quantitative and qualitative simulation experiment results are given to confirm the performance of the proposed method. Table I provides a quantitative performance comparison of the various methods in 4× enlargement for test sets I and II. Visual comparisons of some of the SR results are provided in Figs. 6–8 to perceptually compare the performance of the proposed method, and the ground-truth HR images are given for reference.

A comparison of the average PSNR and SSIM results for each scene class is provided in Table I. It can be seen that the proposed method achieves the highest PSNR and SSIM values. As is well known, with the significant progress of deep learning, the deep learning based video super-resolution (VSR) approaches have greatly improved the performance on synthetic LR frames. Compared to the state-of-the-art VSR method (BasicVSR), the proposed method achieves a better performance and is slightly superior on average for both test sets I and II, which shows the generalization ability of the model. Based on the overall results, these two methods are then followed by L0RIG, RBPN, and SWBTV. The model-based methods (L0RIG and SWBTV) achieve superior results in the remote sensing image SR, due to the effective priors as well as the global reconstruction constraint. It should be noted that some classes, such as baseball diamond, beach, and golf course, present a relatively high PSNR due to the fact that the images associated with these classes are characterized by a greater degree of smoothness than those of the other classes. Among these categories, images in the harbor category have the lowest PSNR of 20.179 dB with the proposed method, which is still better than the other methods. Since texture details and sharp edges are essential in all ground features, the proposed method with the combined constraint achieves notable SR performance improvements for the remote sensing images in each class.

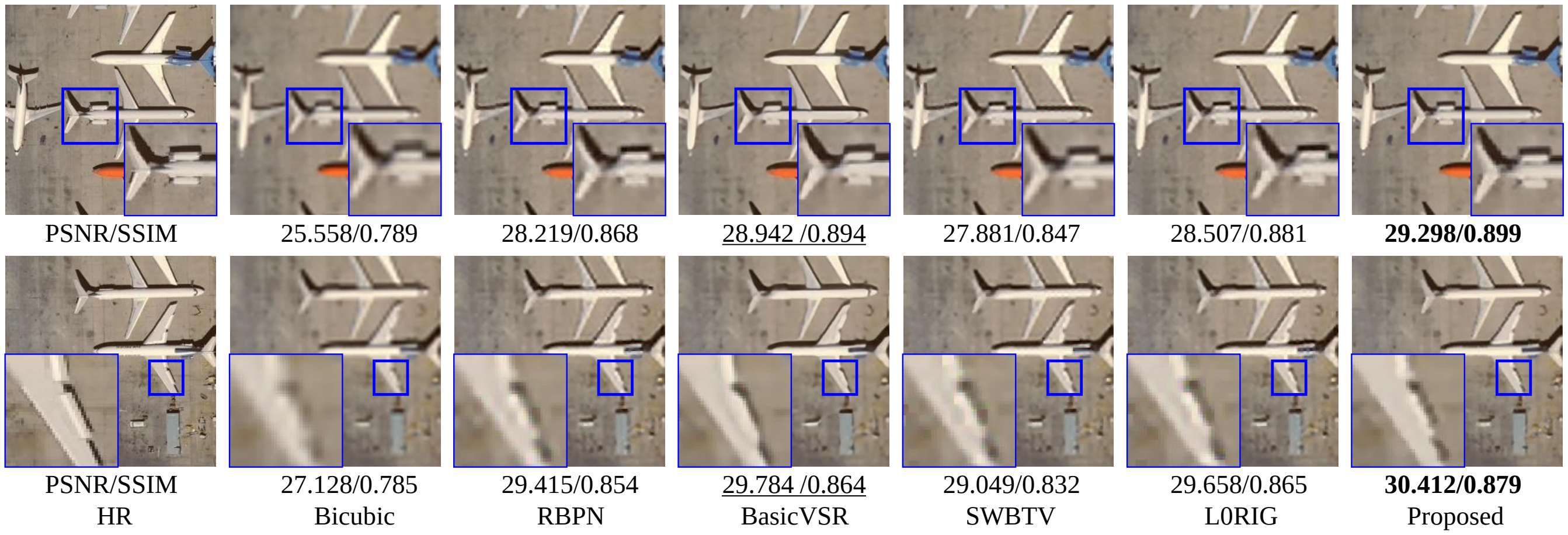

Fig. 6. Visual results for “airplane90” and “airplane91” in test set I using the various methods in 4× SR.

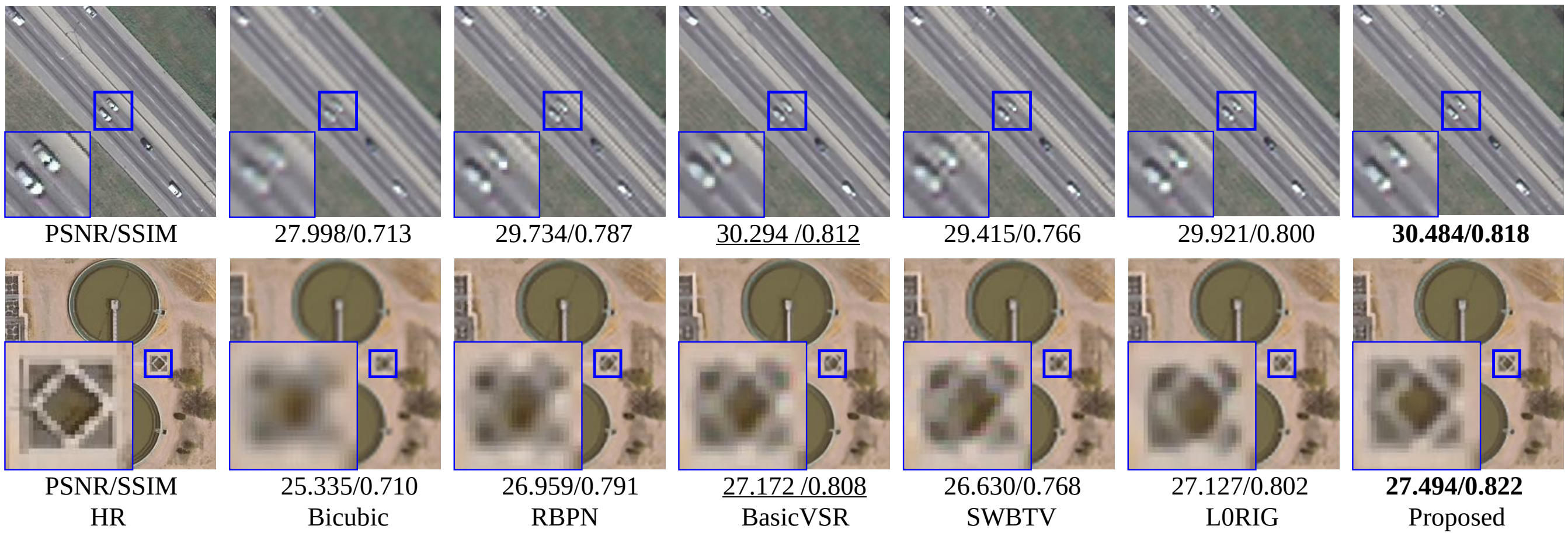

Fig. 7. Visual results for “freeway23” and “storagetanks98” in test set I using the various methods in 4× SR.

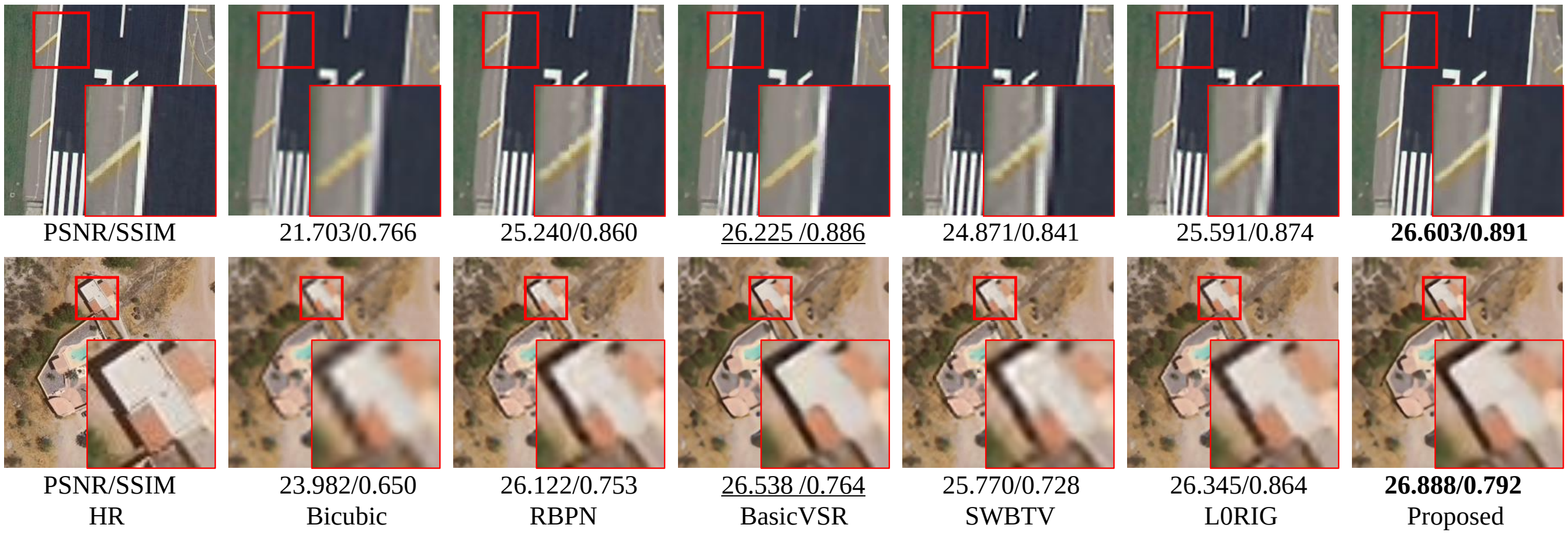

Fig. 8. Visual results for “runway50” and “sparseresidential92” in test set II using the various methods in 4× SR.

TABLE I
AVERAGE PSNR (dB) AND SSIM RESULTS OF EACH CLASS IN 4×SR

| Data | Bicubic | Learning-based | | Model-based | | |
|---|---|---|---|---|---|---|
| | | RBPN [48] | BasicVSR [49] | SWBTV [50] | L0RIG [51] | Proposed |
| Airplane | 25.815/0.751 | 28.439/0.832 | 28.803/0.857 | 28.014/0.822 | 28.737/0.840 | **28.981/0.872** |
| Baseball diamond | 30.358/0.778 | 32.397/0.829 | 32.775/0.844 | 32.028/0.821 | 32.605/0.836 | **32.958/0.861** |
| Buildings | 21.461/0.661 | 24.115/0.795 | 24.484/0.814 | 3.703/0.782 | 24.337/0.807 | **24.632/0.827** |
| Dense residential | 3.355/0.672 | 26.002/0.795 | 26.308/0.818 | 25.575/0.784 | 26.179/0.802 | **26.475/0.834** |
| Freeway | 26.207/0.675 | 28.355/0.777 | 28.732/0.798 | 28.069/0.773 | 28.593/0.789 | **28.965/0.812** |
| Harbor | 17.379/0.691 | 19.927/0.854 | 20.108/0.870 | 19.835/0.848 | 20.097/0.867 | **20.179/0.883** |
| Intersection | 24.343/0.692 | 26.178/0.790 | 26.601/0.801 | 25.966/0.781 | 26.466/0.795 | **26.737/0.816** |
| Mobile home park | 21.841/0.663 | 24.869/0.794 | 25.135/0.805 | 24.407/0.783 | 24.978/0.799 | **25.292/0.822** |
| Storage tanks | 23.549/0.698 | 25.685/0.795 | 26.043/0.815 | 25.388/0.790 | 25.955/0.804 | **26.265/0.825** |
| Tennis court | 28.073/0.780 | 30.420/0.857 | 30.801/0.881 | 30.114/0.852 | 30.588/0.865 | **30.998/0.894** |
| **Average test I** | 24.238/0.706 | 26.639/0.812 | 26.979/0.830 | 26.310/0.804 | 26.854/0.820 | **27.148/0.845** |
| Agricultural | 25.169/0.433 | 25.993/0.493 | **26.210/0.510** | 25.806/0.485 | 26.055/0.495 | 26.196/0.509 |
| Beach | 33.145/0.839 | 34.436/0.876 | 34.797/0.887 | 34.205/0.870 | 34.668/0.879 | **34.964/0.896** |
| Chaparral | 24.082/0.646 | 25.632/0.742 | 26.053/**0.768** | 25.292/0.735 | 25.901/0.751 | **26.102**/0.763 |
| Forest | 26.267/0.603 | 27.177/0.679 | 27.534/0.694 | 26.946/0.672 | 27.421/0.687 | **27.703/0.705** |
| Golf course | 30.718/0.769 | 32.380/0.815 | 32.745/0.827 | 32.012/0.807 | 32.556/0.823 | **32.875/0.835** |
| Medium residential | 23.462/0.653 | 25.909/0.765 | 26.204/0.781 | 25.500/0.755 | 26.081/0.771 | **26.397/0.793** |
| Overpass | 23.124/0.642 | 25.716/0.759 | 26.260/0.780 | 25.462/0.746 | 26.009/0.765 | **26.346/0.791** |
| Parking lot | 19.168/0.607 | 20.782/0.737 | 21.149/0.779 | 20.403/0.727 | 20.962/0.762 | **21.320/0.793** |
| River | 26.473/0.658 | 27.499/0.726 | 27.841/0.741 | 27.287/0.714 | 27.745/0.731 | **27.968/0.755** |
| Runway | 26.556/0.719 | 29.896/0.806 | 30.515/0.835 | 29.735/0.798 | 30.380/0.819 | **30.733/0.846** |
| Sparse residential | 26.158/0.671 | 28.018/0.749 | 28.389/0.769 | 27.766/0.742 | 28.285/0.755 | **28.552/0.782** |
| **Average test II** | 25.847/0.658 | 27.585/0.741 | 27.972/0.761 | 27.310/0.732 | 27.824/0.749 | **28.114/0.771** |

To compare the visual qualities, some of the SR results for the test set with a scale factor of 4× are presented in Figs. 6–8. For a better comparison from a subjective perspective, the rectangles show zoomed regions of the corresponding images to compare the qualitative performance of the different methods. The visual effect basically agrees with the objective results. The experimental results show that the HR remote sensing images recovered by the proposed method exhibit sharper edges and clearer contours, compared with the state-of-the-art SR methods. This is consistent with the proposed local and non-local regularization terms, which are capable of providing the HR gradient information and non-local similar textures for the SR reconstruction model.

Specifically, Fig. 6 gives the reconstructed HR images for the "airplane90" and "airplane91" images generated by the various methods. The reconstructed images of the proposed method have a superior visual quality, as the intricate details in the original images are well preserved. In addition, the close-up images illustrate that the proposed method effectively reconstructs the clear wing edge, whereas the other algorithms suffer from varying degrees of blurring and displeasing artifacts. From the visual comparison in Fig. 7, the proposed algorithm reconstructs more accurate results and preserves finer texture details in the super-resolved image, such as the windscreen of the car. In comparison to the other algorithms, the proposed method demonstrates a superior visual performance, effectively alleviating the blurring artifacts and recovering more details in a manner that is more faithful to the ground truth. Given that the images in test set II are quite distinct from those in the training set, this dataset can be used to verify the generalization ability of the proposed method.

Fig. 8 presents the recovered images in 4× enlargement for the "runway50" and "sparseresidential92" images in test set II obtained by the various methods. It can be observed that the proposed method is superior in its ability to generate sharper edges, as evidenced by the traffic index line on the right side of the road. In contrast, the reconstruction artifacts and unnatural structures are still noticeable in the outcomes of the afore-mentioned comparison methods, particularly in the edge regions. In fact, the performance of the proposed method is in consonance with our initial hypothesis: the NLTV regularization term is apt in preserving edge smoothness and suppressing visual artifacts, while the LGR constraint is beneficial in recovering sharp edges and fine structures.

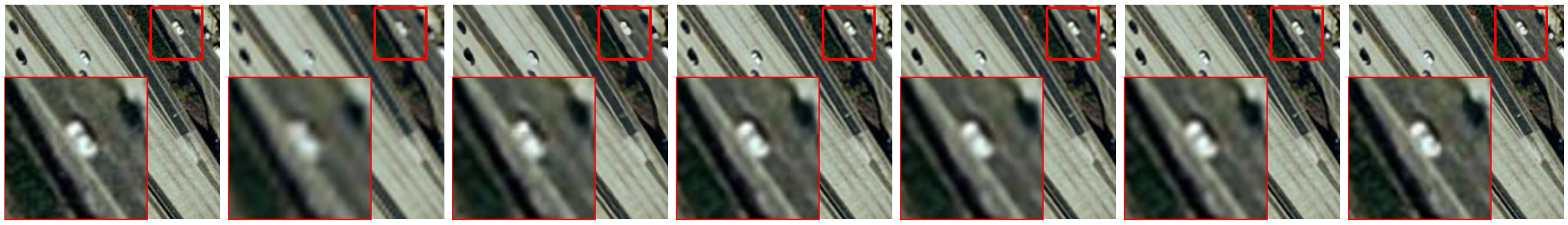

| HR | Bicubic | RBPN | BasicVSR | SWBTV | L0RIG | Proposed |
|---|---|---|---|---|---|---|
| PSNR/SSIM | 27.022/0.770 | 29.647/0.863 | 30.081/0.882 | 29.911/0.875 | 30.226 /0.894 | **30.407/0.901** |

Fig. 9. Visual results for "freeway00" in test set I under Gaussian noise in 4× SR.

TABLE II
AVERAGE PSNR (dB) AND SSIM RESULTS FOR A NOISY IMAGE IN 4×SR

| Metric | Bicubic | RBPN [48] | BasicVSR [49] | SWBTV [50] | L0RIG [51] | Proposed |
|---|---|---|---|---|---|---|
| PSNR | 24.028 | 24.893 | 25.257 | 25.029 | 25.426 | **25.934** |
| SSIM | 0.686 | 0.749 | 0.776 | 0.763 | 0.792 | **0.807** |

Overall, the proposed method achieves the best performance among all the methods in terms of PSNR/SSIM values and visual quality in remote sensing image SR. The average PSNR/SSIM gains on the whole test set over the RBPN, BasicVSR, SWBTV, and L0RIG methods are 0.519 dB/0.032, 0.155 dB/0.012, 0.821 dB/0.040, and 0.292 dB/0.023, respectively. The experimental results demonstrate that the proposed method, which combines the constraints of LGR and NLTV, allows for the reliable recovery of high-frequency details while preserving strong edges and contours with minimal artifacts.

### *C. Robustness Analysis*

Since image noise makes the SR problem more challenging, we analyzed and justified the robustness of the proposed method with regard to noise. Extra additive white Gaussian noise of variance 0.005 was added to corrupt the LR images in test set I. The other parameters were the same as in the previous experiments. Due to the limited space, we only report the average PSNR/SSIM scores of each method in Table II. For the visual quality comparison, the SR results for the noisy "freeway00" image are presented in Fig. 9. Details of the output reconstructed images are given for better illustration.

In terms of the objective performance, the proposed method outperforms the others in terms of PSNR and SSIM results in the quantitative evaluation. Generally speaking, learning-based approaches are susceptible to noise, resulting in a significant drop in their performance on noisy images. Specifically, in the experiment with a noise variance of 0.005, the average PSNR result of the proposed method is 25.934 dB, which is better than that of RBPN, BasicVSR, SWBTV, and L0RIG, by 1.041 dB, 0.677 dB, 0.905 dB, and 0.508 dB, respectively. Fig. 9 illustrates the comparative performances of the various methods in an enlarged area within the red boxes. It can be observed that the comparison learning-based methods fail to effectively suppress the noise artifacts. While RBPN also reproduces fine details well, it introduces unpleasant noise artifacts. The SWBTV method based on bilateral total variation partially suppresses image noise and preserves edges, but exhibits visual artifacts around the edge regions. As the partial enlargement shows, L0RIG demonstrates an optimal balance between noise removal and edge preservation, but fails to recover the lost fine details. The BasicVSR algorithm produces undesirable edge artifacts, including the appearance of artificial edges on flat surfaces and an inadequate suppression of image noise in the textured areas. In contrast, the proposed method demonstrates a superior performance, exhibiting clear high-frequency details and fewer ringing artifacts, as shown in Fig. 9. It is noteworthy that the distorted content, such as the windscreen of the car, can be finely restored with the proposed approach. Compared to the other methods evaluated on noisy images, the proposed method performs favorably due to its learned gradient constraint, which enhances edge sharpness while suppressing jaggy artifacts along high-frequency structures.

In conclusion, based on both the qualitative and quantitative analyses, the results consistently demonstrate that the proposed method exhibits a superior SR performance in accurately recovering high-frequency information that closely resembles the ground-truth image. By leveraging a combined prior, the proposed approach effectively enhances edge sharpness and preserves the intricate texture details of the recovered image. Consequently, it significantly outperforms the comparison methods by a substantial margin in terms of PSNR, thereby further validating its superiority in reconstructing textures and enhancing edges. Moreover, the comparative evaluation against other state-of-the-art SR approaches confirms the robustness and effectiveness of the proposed method in terms of noise corruption.

### *D. Real-Data Experiments*

In this section, we describe how we employed real sequences of remote sensing images to further evaluate the robustness of the proposed method. The three sequence of real LR remote sensing images were extracted from the Jilin No. 1 satellite video dataset [1], each consisting of 16 consecutive frames with a fixed size of 256 × 256. The resolution of the Jilin-1 imagery (1.12 m/pixel) is within the resolution range of our test set, which is 1.2 m/pixel (0.3 × 4) for a scale factor of 4. Since no ground-truth HR image was available for the experiments on the real sequences, the no-reference image evaluation metrics of NIQE [53] and PIQE [54] were employed to further evaluate the quantitative quality of the real image sequence SR results. Visual comparisons of the results for the Jilin-1 satellite imagery obtained by the different methods are shown in Fig. 10 and Fig. 11. In each image, we have marked a local region of interest and displayed its zoomed version separately alongside the image.

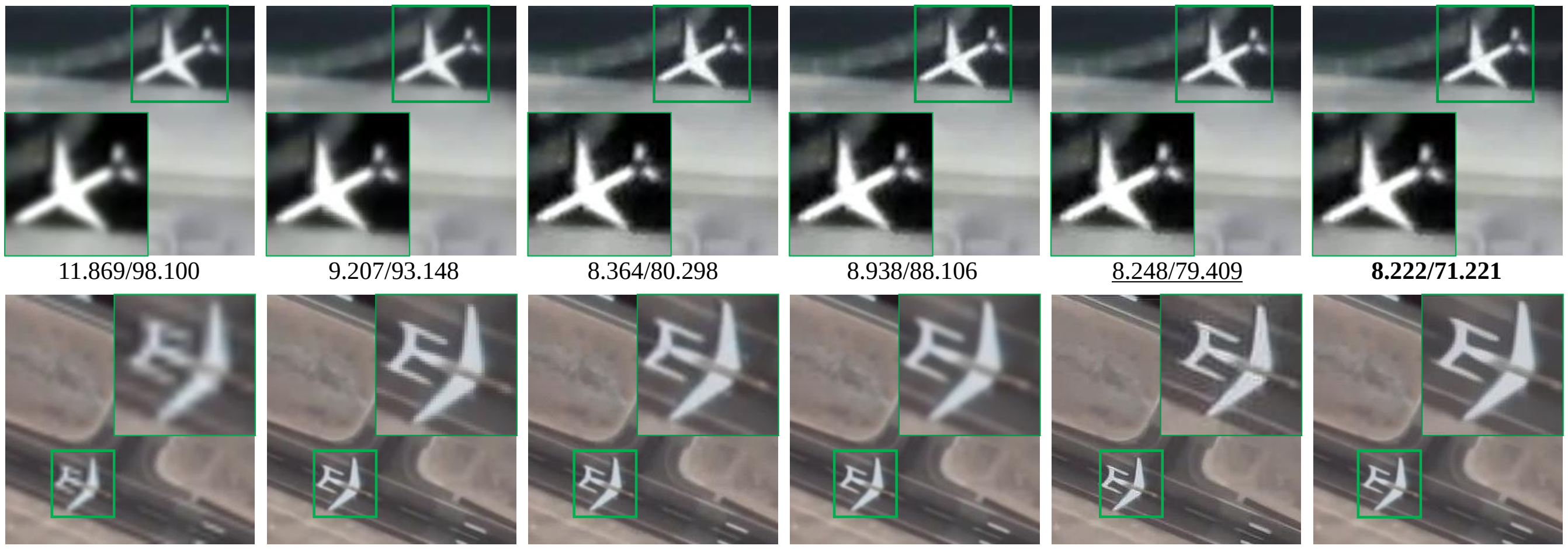

| Bicubic | RBPN | BasicVSR | SWBTV | L0RIG | Proposed |
|---|---|---|---|---|---|
| 11.869/98.100 | 9.207/93.148 | 8.364/80.298 | 8.938/88.106 | 8.248/79.409 | **8.222/71.221** |
| 11.382/90.942 | 9.446/82.586 | 8.834/68.360 | 9.148/76.797 | 7.918 /57.327 | **7.896 /46.903** |

Fig. 10. Visual results for the "airplane" scene in the Jilin-1 satellite imagery obtained by the different methods in 4× SR, where the numbers below refer to the NIQE and PIQE values of the corresponding results.

The experimental results on the real image sequences show that the proposed method yields a superior performance in both the objective metrics and visual quality. For a real-world image, the down-sampling kernel is typically unknown and complicated, which significantly affects the performance of the non-blind SR methods. Nevertheless, the proposed method can produce visually pleasant images and effectively suppress the errors caused by noise, registration, and poor estimation of unknown point spread function (PSF) kernels.

In the case of the "airplane" scene depicted in Fig. 10, the proposed method generates superior results with a diminished prevalence of jagged and ringing artifacts, particularly in instances where motion estimation errors are minimal. The recurrent neural network and explicit motion estimation enable the RBPN method to fuse the multi-frame information, which shows its great ability for satellite VSR, as well as the BasicVSR method. The learning-based method shows enhanced edge sharpness for manufactured objects, but also generates false textures and undesired visual artifacts in the SR results. The model-based methods (L0RIG and SWBTV) achieve superior results in the real satellite VSR, due to the effective priors as well as the global reconstruction constraint. While SWBTV is effective at noise suppression, it is obviously inferior to the proposed method in terms of detail preservation and sharp edge recovery. For the "runway" scene in Fig. 11, due to the serious loss of object information and the lack of any reasonable reference objects, all the comparison SR methods show limitations for the small static objects. From the visual results, it can be seen that better edge regions and texture regions are reconstructed by the proposed method, compared with the state-of-the-art SR methods. In contrast, BasicVSR tends to generate more artifacts at the edges, and the result of RBPN suffers from visible ghosting artifacts and is seriously affected by the blur effects.

In summary, the proposed method can improve the reconstruction results for Jilin-1 satellite imagery greatly, in both the visual effects and quantitative results. The experimental results demonstrate that the proposed method is superior to many of the learning-based and model-based MFSR methods, especially in the edge and texture regions. These results also indicate that the proposed model is more robust than the other methods.

### *E. Effectiveness of the Regularization Terms in the Proposed Method*

The proposed method incorporates two regularization terms: the LGR and NLTV priors. Moreover, the LGR prior uses the learned gradient and HR profile transformation to respectively address the issues of edge ambiguity and noise sensitivity. To validate the effectiveness of both the local and non-local regularization terms in the proposed framework, ablation experiments were performed on the images in test set I, with three extended versions of the proposed method. The three extended versions were as follows: NLTV (only the NLTV prior was used), NLTV-LG (the gradient regularization term was constructed with the DRAN), and NLTV-GPT (the gradient regularization term was constructed with the gradient profile transformation). The proposed framework—NLTV-LGR—incorporates both the NLTV and LGR priors, where the LGR prior is constructed with both a gradient-learning network and gradient profile transformation. The bicubic interpolation method was taken as a benchmark for comparison. The average PSNR and SSIM scores for the different extended versions are listed in Table III**.** It can be observed that the complete version (NLTV-LGR) achieves the best performance, indicating that the combination of NLTV and LGR priors is significantly beneficial in enhancing the quality of super-resolved images. Moreover, for the perceptual quality comparison, the super-resolved results of the different extended versions for the "tenniscort28" test image are shown in Fig. 12.

*1) Effectiveness of the Gradient-Learning Network and Gradient Profile Transformation Schemes:* In the proposed

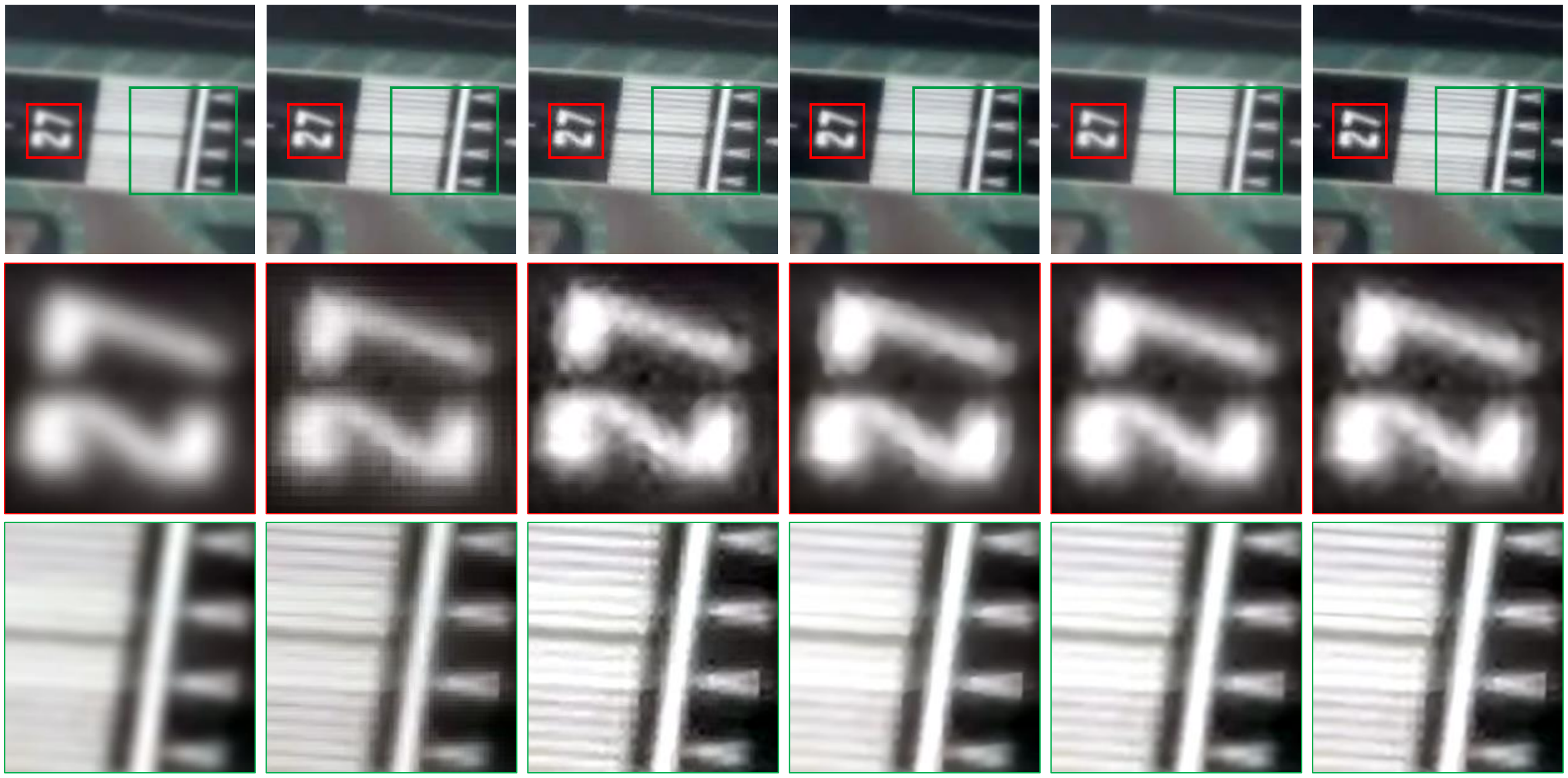

10.466/89.502 Bicubic | 8.787/88.923 RBPN | 7.950/84.355 BasicVSR | 8.103/86.884 SWBTV | 7.849 /83.846 L0RIG | **7.601 /79.098** Proposed

Fig. 11. Visual results for the "runway" scene in the Jilin-1 satellite imagery obtained by the different methods in 4× SR, where the numbers below refer to the NIQE and PIQE values of the corresponding results.

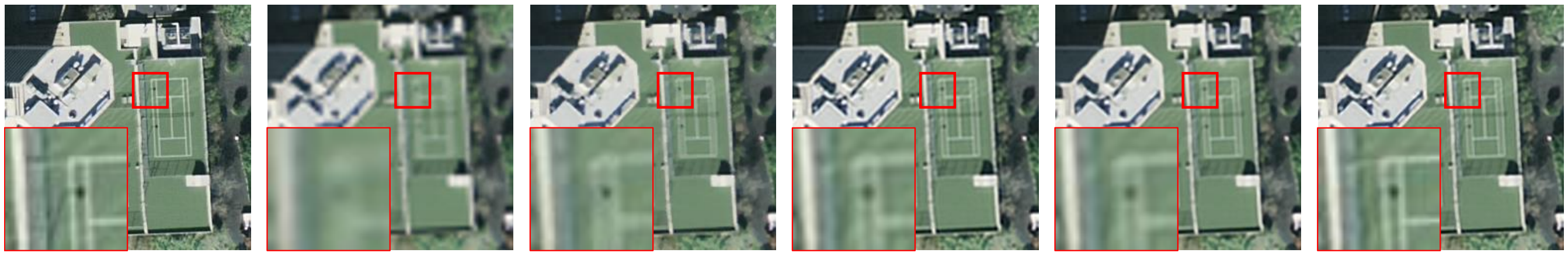

HR PSNR/SSIM | Bicubic 23.018/0.632 | NLTV 26.291/0.767 | NLTV-LG 26.926/0.786 | NLTV-GPT 26.672/0.775 | NLTV-LGR (Proposed) **27.258/0.829**

Fig. 12. Visual results for "tenniscort28" with different extended versions of the proposed method in 4× SR.

method, the estimated HR gradient field is taken as the prior term to constrain the SR reconstruction process. The gradient regularization term can be constructed by either a gradient-learning network (denoted by LG) or gradient profile transformation (denoted by GPT). A comparison was made between NLTV-LG and NLTV-GPT with the original NLTV prior to verify the effectiveness of the LG and GPT schemes. According to the results presented in Table III and Fig. 12, NLTV-LG and NLTV-GPT can produce sharper edges than NLTV. The experimental results indicate that both the developed gradient-learning network and the gradient profile transformation have their own respective advantages. However, as a whole, the gradient-learning network term is more effective than the gradient profile transformation.

*2) Effectiveness of the LGR Prior:* Compared with the other two gradient regularization terms (LG and GPT), the recovered high-frequency details and edges of the proposed NLTV-LGR are more accurate and much clearer. For instance, in the lower-left zoomed regions in Fig. 12, the edges in the results of NLTV-LG and NLTV-GPT are severely blurred or distorted, whereas the edges of the proposed method are much clearer and more similar to the original HR edges. This comparison also verifies the complementary properties of the gradient-learning network and gradient profile transformation. While their respective contributions differ, these components are mutually reinforcing. The quantitative and qualitative results demonstrate that the estimated LGR prior can assist in the generation of visually pleasing SR results with clearer textures for remote sensing imagery.

TABLE III
AVERAGE PSNR (dB) AND SSIM RESULTS OF THE DIFFERENT EXTENDED VERSIONS IN 4×SR

| Method | Bicubic | NLTV | NLTV-LG | NLTV-GPT | NLTV-LGR |
|---|---|---|---|---|---|
| NLTV | × | √ | √ | √ | √ |
| LG | × | × | √ | × | √ |
| GPT | × | × | × | √ | √ |
| PSNR | 24.238 | 26.706 | 26.954 | 26.881 | **27.148** |
| SSIM | 0.706 | 0.821 | 0.831 | 0.825 | **0.845** |

*3) Effectiveness of Combining NLTV with LGR:* Fig. 13 illustrates the average PSNR gains of the proposed NLTV-LGR method over the three other extended versions for each image category in test set I. By coupling the NLTV and LGR priors in the proposed framework, the proposed method achieves a 0.442 dB/0.024 improvement over the use of a separate NLTV prior. With regard to the visual quality, the blur effects remain evident in the result obtained with only the NLTV prior, especially along the edges. In contrast, the LGR prior has advantages in reconstructing accurate edge and texture details. Consequently, the combined prior can produce sharp edges and fine structures, while simultaneously suppressing the reconstruction artifacts well. This demonstrates that the proposed method can fully exploit the complementary advantages of the two priors and can achieve more reliable super-resolved images with fewer artifacts and sharper edges.

TABLE IV
AVERAGE RUNNING TIME (S) ON TEST SET I IN 4×SR

| Method | RBPN [48] | BasicVSR [49] | SWBTV [50] | L0RIG [51] | NLTV | NLTV-LG | NLTV-GPT | Proposed |
|---|---|---|---|---|---|---|---|---|
| Time (s) | 18.953 | 5.799 | 46.583 | 58.832 | 77.782 | 84.065 | 129.922 | 147.917 |

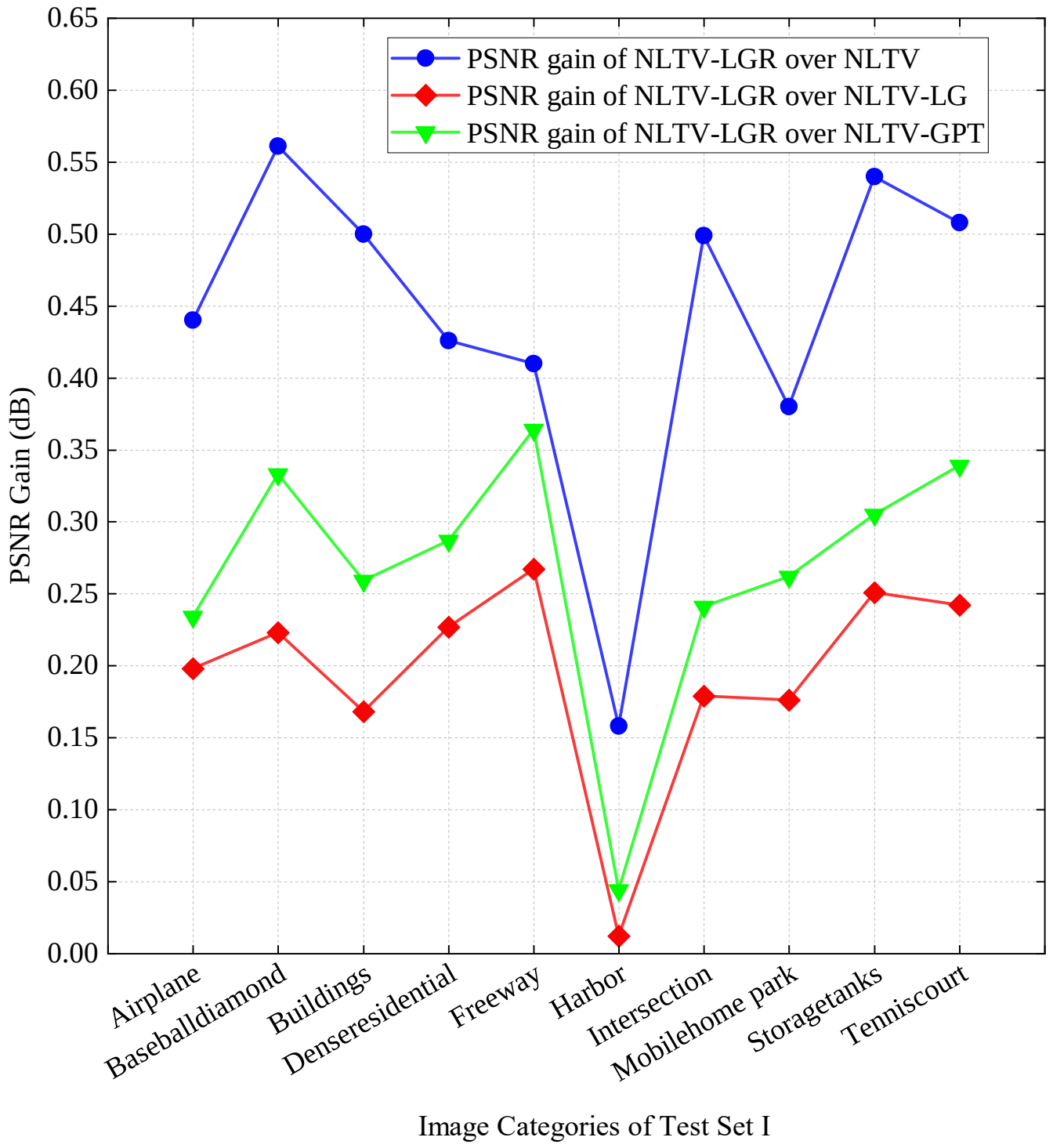

Fig. 13. Comparison of the SR performance between different extended versions for each image category in test set I.

### *F. Empirical Study on Regularization Parameters*

There are two main regularization parameters $\alpha$ and $\beta$ that are crucially important to the SR performance of the proposed method. The selection of regularization parameter has always been a headache for image SR. To make a fair comparison between the different methods, in the case of a small volume of experimental data, the regularization parameters can be determined manually by trying a series of values and selecting the ones with the highest PSNR or the best visual effect. In the experiments conducted in this study, if the regularization parameters had been selected manually, it would have been very time-consuming and tedious since there were more than 1000 images in the test set. In order to set the gradient regularization parameter $\alpha$ and the NLTV regularization parameter $\beta$ reasonably, we adopted the adaptive parameter selection method [55] to obtain the approximate optimal regularization parameters in most cases of the simulation experiments.

### *G. Discussion on Computational Cost*

By roughly analyzing the proposed method, the following four parts are the main computational cost: 1) estimation of the guidance gradient field with gradient profile transformation; 2) parameter estimation of the motion and blur kernel; 3) similar patch search for the NLTV prior; and 4) inner iterative optimization with ADMM.

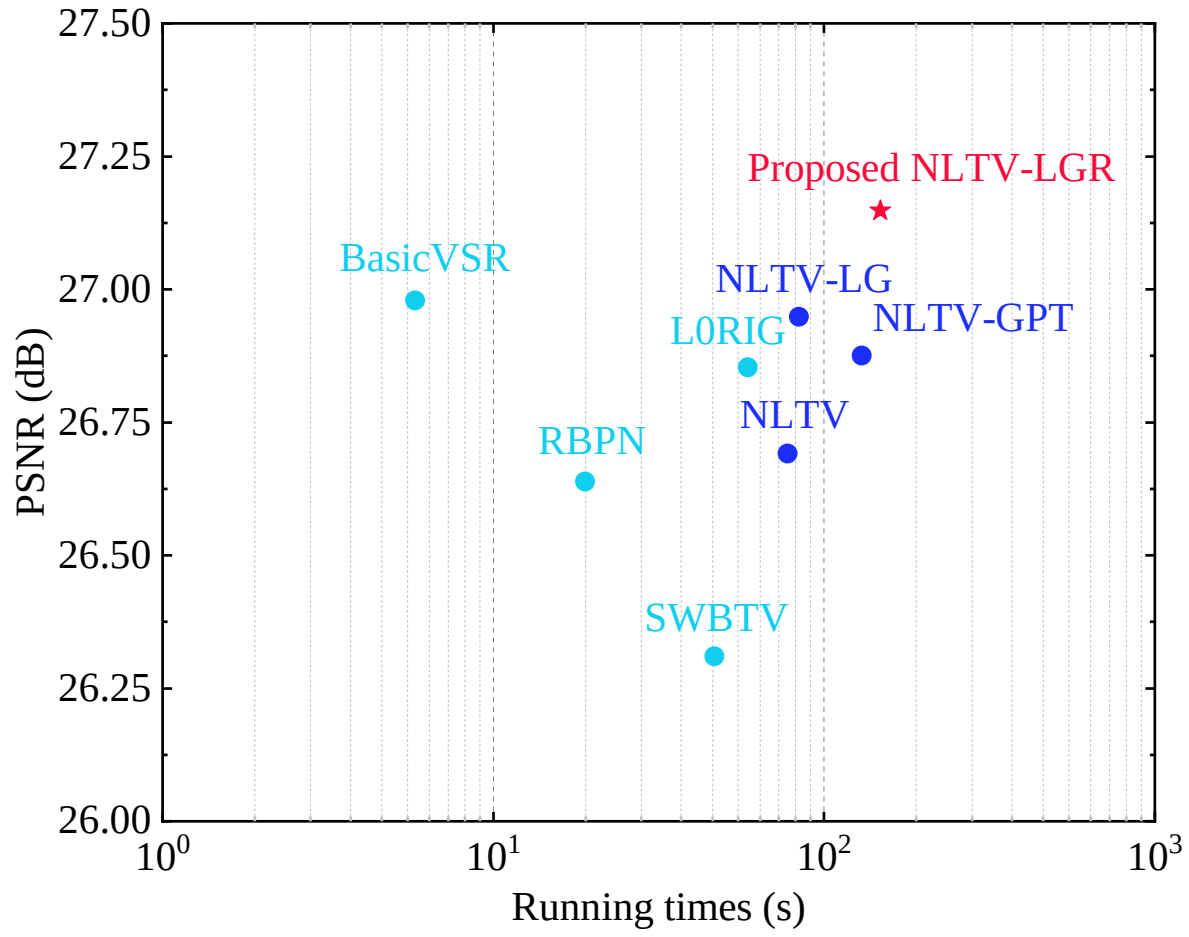

Fig. 14. Average PSNR scores versus running times on test set I.

To comprehensively evaluate the computational cost of the proposed method, the average PSNR values versus running times of the proposed method and the other comparison methods on the images of test set I are shown in Fig. 14. All the experiments were conducted with MATLAB R2022b on an Intel Core i7 3.5 GHz CPU. Owing to their inner iterative optimization process, the model-based methods are more time-consuming than the learning-based methods, as illustrated in Table IV. In contrast, the deep learning based methods are extremely fast in the SR process due to their natural ability for massively parallel computing. However, in this comparison, the complex and time-consuming training process of the learning-based methods has not been taken into account. Specifically, the average running time of the proposed method is approximately 147.917 s, which is considerably slower than that of the comparison methods. Therefore, the computational cost is indeed a drawback of the proposed approach, although it significantly enhances the SR performance for remote sensing images.

## IV. Conclusions

In this paper, we have introduced a framework that combines the advantages of both deep learning based and variational model based approaches. The gradient-guided MFSR method based on NLTV and LGR priors simultaneously exploits the inter-frame aliasing information and external feature learning. Specifically, the DRAN is employed to learn the latent HR gradient, which is then constructed as a local reconstruction constraint through the gradient profile transformation for recovering fine image details. The well-designed network is capable of mining high-frequency spatial information that can be used to assist in restoring sharp edges in the HR image. Meanwhile, the NLTV prior is incorporated into the proposed MFSR framework as a global regularization term, which can effectively utilize the non-local spatial similarity within remote sensing imagery, thereby enhancing the texture details and suppressing image artifacts. By combining the two complementary priors within the adaptive norm based reconstruction framework, the mixed $L1$ and $L2$ regularization minimization problem is solved via an iterative optimization algorithm.

Both the simulation and real-data experiments on remote sensing images illustrated that the proposed framework works more effectively than the other comparison SR methods, especially in the restoration of fine details. The proposed method can achieve superior evaluation metrics and visual results with higher fidelity and richer high-frequency information. In addition, we have demonstrated that the combination of the local and non-local regularization terms can further enhance the SR performance by imposing a local texture detail and global structure information constraint on remote sensing SR. In our future work, we will further examine the coupling of the variational model based and the deep learning based MFSR methods in order to bring out their respective advantages.